%% file: paper.tex
\pdfoutput=1

\documentclass[11pt]{article}
\makeatletter
\renewcommand*{\@opargbegintheorem}[3]{\trivlist
      \item[\hskip \labelsep{\bfseries #1\ #2}] \textbf{(#3)}\ \itshape}
\makeatother
\usepackage{listings}
\usepackage[final]{Style/acl}
\usepackage{float}
\usepackage{times}
\usepackage{latexsym}
\usepackage{kotex}
\usepackage{natbib}
\usepackage{algorithm}
\usepackage{algpseudocode}
\usepackage{algorithmicx} 
\usepackage{float}
\usepackage{wrapfig}
\usepackage{graphicx}
\usepackage{subcaption}
\usepackage{booktabs} 
\usepackage{adjustbox}
\usepackage{multirow}
\usepackage{float}
\newtheorem{define}{Definition}

\newtheorem{theorem}{Theorem}

\usepackage{Style/math_commands}
\usepackage{subcaption}
\usepackage{comment}
\algnewcommand{\LineComment}[1]{\State \textcolor{blue}{\texttt{// #1}}}

\usepackage[T1]{fontenc}

\usepackage[utf8]{inputenc}

\usepackage{microtype}

\usepackage{inconsolata}

\usepackage{graphicx}

\title{Public Data Assisted Differentially Private In-Context Learning}

\author{Seongho Joo \\
  Seoul National University \\
  \texttt{seonghojoo@snu.ac.kr} \\ \And
  Hyukhun Koh \\ 
  Seoul National University \\ 
  \texttt{hyukhunkoh-ai@snu.ac.kr} \\ \And
  Kyomin Jung\footnotemark[2] \\
  Seoul National University  \\
  \texttt{kjung@snu.ac.kr} \\}

\begin{document}
\maketitle
\begingroup
\renewcommand\thefootnote{\dag}  
\footnotetext[2]{Corresponding author.}
\endgroup
\input{Sections/abstract}
\input{Sections/introduction}
\input{Sections/relatedworks}
\input{Sections/preliminaries}
\input{Sections/DP-ICL}
\input{Sections/experiments}

\input{Sections/Analysis}
\input{Sections/conclusion}
\input{Sections/limitation}
\bibliography{Sections/reference}
\input{Sections/appendix}

\end{document}

%% file: Sections/abstract.tex
\begin{abstract}
In-context learning (ICL) in Large Language Models (LLMs) has shown remarkable performance across various tasks without requiring fine-tuning. However, recent studies have highlighted the risk of private data leakage through the prompt in ICL, especially when LLMs are exposed to malicious attacks. While differential privacy (DP) provides strong privacy guarantees, it often significantly reduces the utility of in-context learning (ICL). To address this challenge, we incorporate task-related public data into the ICL framework while maintaining the DP guarantee. Based on this approach, we propose a private in-context learning algorithm that effectively balances privacy protection and model utility. Through experiments, we demonstrate that our approach significantly improves the utility of private ICL with the assistance of public data. Additionally, we show that our method is robust against membership inference attacks, demonstrating empirical privacy protection.
\end{abstract}

%% file: Sections/introduction.tex
\section{Introduction}
With the emergence of Large Language Models (LLMs), in-context learning (ICL) has demonstrated remarkable performance across various tasks by enabling models to infer from provided examples without modifying internal parameters \cite{Brown,Min2022RethinkingTR,Wei2022EmergentAO}. This flexibility allows LLMs to adapt to diverse domains without explicit training, leading to its widespread adoption \cite{Dong2022ASO}.

\begin{figure}[t]
  \includegraphics[width=\columnwidth]{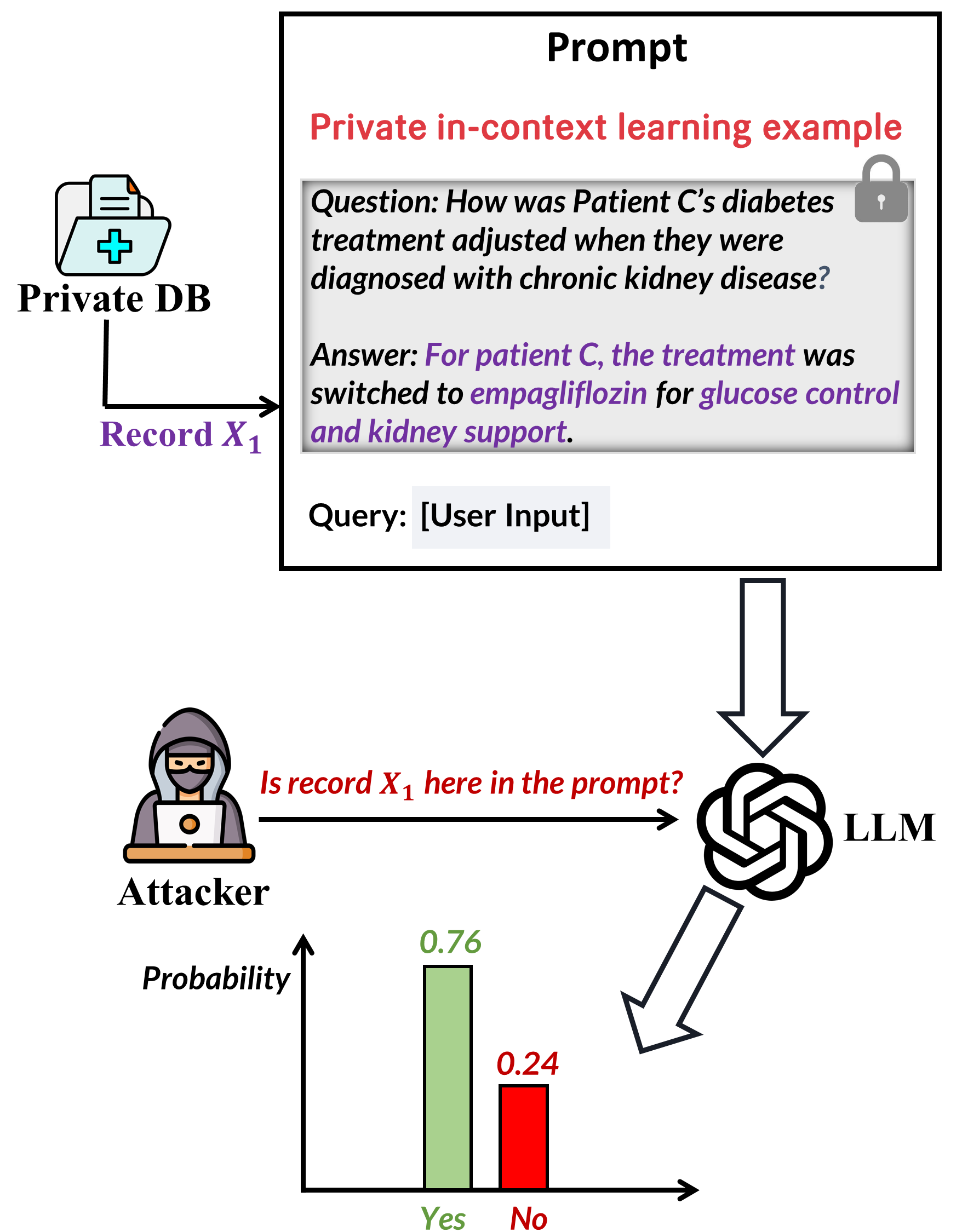}
    \caption{An illustration of a \emph{membership inference attack}, where a malicious attacker infers whether a target example is included in the prompt composed of private examples retrieved from the database. By leveraging the tuned prompt, the attacker can determine the presence of the target example within the in-context examples.}
  \label{fig:attackexample}
  \vspace{-0.5cm}
\end{figure}

Meanwhile, concerns about privacy leakage have been raised regarding the deployment of ICL in LLMs \cite{Li2023PrivacyIL,Kandpal2023BackdoorAF,Li2024ChainofScrutinyDB}. For a concrete example, consider the ICL scenario depicted in Figure \ref{fig:attackexample}, where private patient treatment records are used as demonstration examples. A malicious attacker aiming to identify private clinical record data may try inference attack targeting LLM. The LLM with an ICL algorithm, without privacy protection, could potentially expose sensitive clinical data to attackers. Such exposure of personal health information may contravene data protection regulations like the General Data Protection Regulation (GDPR), which mandates appropriate technical and organizational measures to ensure data security and protect individual privacy rights \cite{GDPR2016}.


Differential privacy (DP) has emerged as the gold standard for rigorous privacy protection across multiple domains, including computer vision, recommendation systems, natural language processing, and census data \cite{Dwork2006DifferentialP}. 
The essence of differential privacy is to ensure that the output of an algorithm is minimally influenced by the inclusion or exclusion of any individual’s data, thereby reducing the risk of privacy leakage. Additionally, the level of privacy protection can be adjusted by tuning the privacy protection parameter $\varepsilon$. Differential privacy is often employed in conjunction with \emph{sample and aggregate} techniques, where multiple noisy outputs are combined to enhance privacy.  
 
However, when DP is integrated into ICL algorithms, two main challenges arise.
First, the performance of the ICL algorithm deteriorates significantly under strong privacy protection, where output perturbation negatively impacts the model's utility. To mitigate this degradation, we incorporate public data into our ICL framework to minimize utility loss.
The second challenge is to design a privacy-preserving aggregation method for language generation outputs. Given the high dimensionality inherent in the output space of LLM responses, it is essential to transform them into a lower-dimensional representation while ensuring that they remain reconstructable in the original response space. To achieve this, we project LLM-generated responses into semantic embeddings and form \emph{semantic groups} for private aggregation.

We evaluate our private ICL framework with DP guarantees on question-answering tasks (ChatDoctor, \citep{li2023chatdoctor}) and a document summarization task (SAMsum, \citep{gliwa-etal-2019-samsum}). The experimental results show that our private ICL framework performs comparably to non-private baselines at a strong privacy protection level ($\varepsilon=1$) and outperforms the private data-only counterpart at the same privacy protection level. We further demonstrate that using out-of-distribution (OOD) public data, as well as in-distribution (ID) public data, is beneficial for minimizing utility degradation. Moreover, we show that our private ICL framework is robust against empirical privacy attacks by simulating membership inference attacks on the model.
In summary, our main contributions are the following:
\begin{enumerate}
\item We propose a DP-guaranteed private ICL framework that integrates public data and employs semantic group aggregation to manage high-dimensional outputs.
\item Our framework demonstrates effective utility-privacy tradeoffs in question answering and summarization tasks, and benefits from both ID and OOD public data.
\item Our approach is robust against empirical privacy threats, as verified by membership inference attack experiments.
\end{enumerate}

%% file: Sections/relatedworks.tex
\begin{figure*}[t]
\begin{center}
  \includegraphics[width=.9\linewidth]{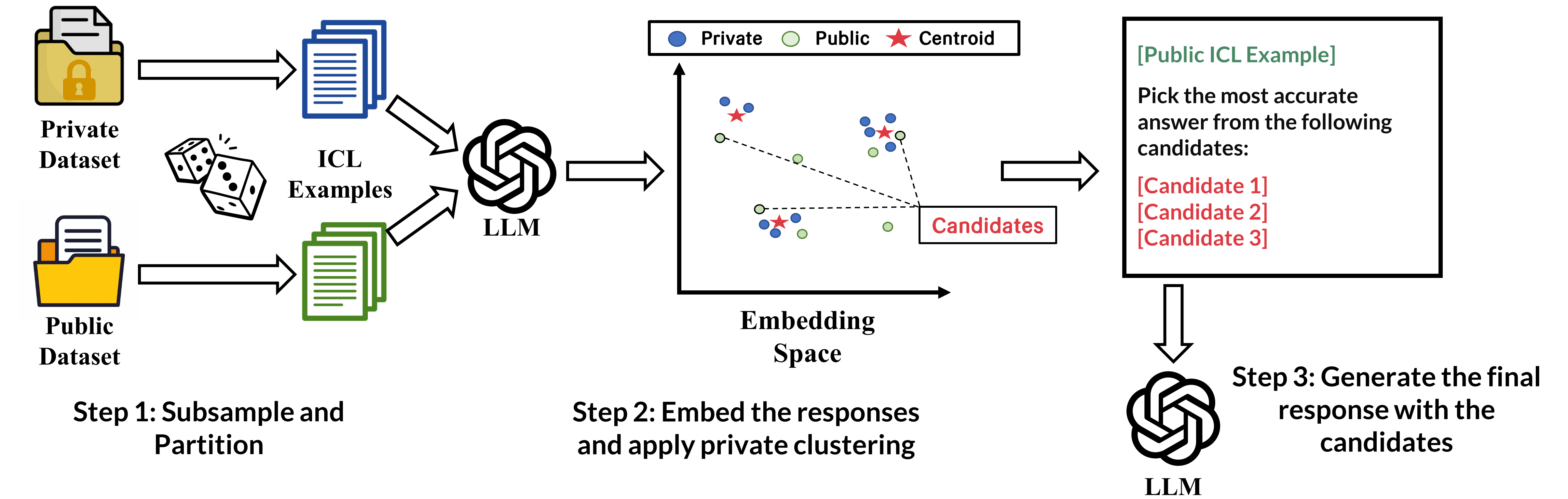} 
  \caption {\textbf{Overview of public data-assisted differentially private in-context learning.} In Step 1, the public and private data are partitioned and used to create demonstration examples. In Step 2, the generated responses of the LLMs are embedded into a semantic space and then clustered privately. Finally, in Step 3, the top-$k$ candidates closest to the centroids are selected, and the LLM chooses the final answer from among them with a public ICL example.}
  \label{fig:model_overview}
\end{center}
  \vspace{-0.8cm}
\end{figure*}
\section{Related Work}
\paragraph{Private Text Generation}
Differentially private text generation methods can be broadly categorized into differentially private fine-tuning approaches \cite{Yu2021DifferentiallyPF,Li2021LargeLM,Yu2023SelectivePF} and differentially private prediction approaches \cite{Majmudar2022DifferentiallyPD,Ginart2022SubmixPP,Flemings2024DifferentiallyPN}. Fine-tuning methods fine-tune LLMs using the DP-SGD algorithm \cite{Abadi2016DeepLW}. However, private fine-tuning methods suffer from high computational costs and cannot be applied to black-box LLMs. Moreover, private fine-tuning does not allow for flexible data replacement.

In prediction methods, many works build upon the PATE framework \cite{Papernot2018ScalablePL}, where each teacher model is trained on a subset of the private dataset, and the aggregate of the teacher ensemble is used for prediction. Apart from methods with differential privacy guarantees, text sanitization methods \cite{Albanese2023TextSB, Papadopoulou2022NeuralTS} and unlearning methods \cite{kassem-etal-2023-preserving} have also been proposed to protect individual privacy.

\paragraph{Differentially Private In-Context Learning}
\citet{Duan2023FlocksOS} propose \texttt{PromptPATE}, where unlabeled public data is privately annotated by a teacher ensemble trained on the private dataset, and the annotated public data is then used as demonstration examples in the prompt for ICL. \citet{Tang2023PrivacyPreservingIL} propose a \emph{token-level} differentially private ICL framework that generates synthetic texts as demonstration examples for subsequent queries. However, the output space of this ICL framework is limited to the label space and does not cover long-text responses. \citet{wu2024privacypreserving} develop an ICL framework with DP guarantees that can handle long-text responses. To address the high-dimensionality issue of the output text space, they propose embedding space aggregation and keyword space aggregation. However, the performance with strong privacy protection is not satisfactory compared to the non-private counterpart.

\paragraph{Differential Privacy with Public Data}
As a seminal work, \citet{Wang2020DifferentiallyPL} addresses an empirical risk minimization problem with limited public data and proposes a private-public stochastic gradient descent (SGD) method that uses public data to adjust training parameters. Subsequently, variants of DP-SGD that leverage public data have demonstrated the efficiency of public data by showing improvements in utility \cite{Nasr2023EffectivelyUP} and have shown that public data can be used to learn informative priors for efficient private learning \cite{Tang2023PrivacyPreservingIL}. While earlier works primarily applied public data during the fine-tuning stage, our approach integrates public data during the inference stage. With careful incorporation of public data, our ICL framework demonstrates a superior privacy-utility trade-off compared to baseline methods.  We present the comparison table from the previous DP literature in Table \ref{tab:comparison_methods} in Appendix. 

%% file: Sections/preliminaries.tex
\section{Preliminaries}
\subsection{In-Context Learning}
To respond to the user query $Q$, the demonstration examples $(Q_1, A_1), \dots, (Q_n, A_n)$ are concatenated with the user query to assist in generation. The LLM can learn from the demonstration examples by identifying the relevant mapping from each $(Q_i, A_i)$ in the examples. Given the query and the demonstrations, the LLM selects the next token using various sampling algorithms or greedy decoding. The objective of our work is to ensure that the algorithm remains private, such that an attacker cannot infer the presence of specific target data within the demonstration examples in in-context learning (ICL).

\subsection{Differential Privacy}
Differential privacy \cite{Dwork2006DifferentialP} is considered the gold standard for protecting the privacy of machine learning algorithms. The formal definition is as follows: \begin{define}[Differential Privacy] A randomized mechanism $\cM: \cD \to O$ is said to satisfy $(\varepsilon, \delta)$-differential privacy if for any neighboring datasets $\cD$ and $\cD^{\prime}$, which differ in only a single element, it holds that \begin{equation*} \Pr[\cM(\cD) \in \cS] \leq e^{\varepsilon} \Pr[\cM(\cD^{\prime}) \in \cS]  + \delta \end{equation*} for any set $\cS$ of possible outputs in $O$. \end{define}

As $\varepsilon$ decreases, the probabilities $\Pr[\cM(\cD) \in \cS]$ and $\Pr[\cM(\cD^{\prime}) \in \cS]$ become closer, making it more difficult for an attacker. The parameter $\delta$ represents the failure probability, where the DP guarantee may not hold.

\paragraph{Post-processing Property}
The post-processing property allows for arbitrary transformations of the output of a DP algorithm, enabling the development of DP algorithms that exploit public data.
\begin{define}[Post-processing of DP]
\label{postprocessing}
Let $\mathcal{M}$ be an $(\varepsilon, \delta)$-differentially private algorithm, and let $f$ be any transformation function. Then $f(\mathcal{M}(D); D_{pub})$ also satisfies $(\varepsilon, \delta)$-differential privacy, where $D_{pub}$ is auxiliary public data.
\end{define}

%% file: Sections/DP-ICL.tex
\section{Private ICL with Public Data}
In this section, we present our private ICL framework. We first explain each stage of the private ICL framework and then present the DP privacy analysis of the framework. The schematic diagram of our private-ICL framework is presented in Figure \ref{fig:model_overview} and the algorithm is described in Algorithm \ref{algo:privateICL}.  

\subsection{Private ICL}
\paragraph{Step 1: Dataset Subsampling and Partition} Before providing the demonstration dataset to the LLMs, we randomly subsample a fraction of $p\%$ from the dataset. Subsampling offers two key advantages: (1) \textbf{Privacy Amplification}: For an $(\varepsilon, \delta)$-DP algorithm $\mathcal{M}$, subsampling can amplify privacy, effectively reducing the privacy loss. Specifically, under certain subsampling techniques, the privacy parameter $\varepsilon$ can be reduced to approximately $p\varepsilon$, where $p$ is the subsampling rate\footnote{For stability of the algorithm, we choose the uniform sampling without replacement. The details of privacy amplification are explained in Theorem \ref{thm:amplification}.}. (2) \textbf{Memory Cost Reduction}: By using a subset of the dataset instead of the entire set of demonstration examples, we can significantly reduce the memory load on the API. In the $n$-shot, $m$-ensemble setting, we uniformly subsample $mn$ demonstration examples from both the private and public datasets. Each $n$-shot example is formatted into a task-specific prompt with the user query and fed to the LLM. 
After inference, we obtain $m$ responses for each example.
\paragraph{Step 2: Private Aggregation of Responses}
After generating multiple responses from the ensemble, we need to aggregate them privately to prevent attackers from inferring information about the private examples. However, direct aggregation results in a highly sparse histogram due to the nearly infinite-dimensional output space of LLMs.

To address this, we adopt \emph{private clustering} from \citet{Li2024ChainofScrutinyDB} (\texttt{DPM} of line 9 in Algorithm \ref{algo:privateICL}). First, we obtain embeddings for each response using a text embedding model. Then, we apply private k-means clustering to both private and public embeddings, generating privatized cluster centers and the number of members (weights) in each cluster.
For class representatives, we select the closest public element to each cluster center (as shown in line 14 of Algorithm \ref{algo:privateICL}).
As a baseline aggregation method, we employ a modified version of Keyword Space Aggregation (KSA) from \cite{wu2024privacypreserving}, in which sentences are projected into a keyword space, and responses are reconstructed using the most frequent keywords.

\begin{algorithm}[h]
\caption{Public-data assisted In-Context Learning}
\label{algo:privateICL}
\footnotesize
\begin{algorithmic}[1]
\Require \textbf{LLM}, private dataset $\mathcal{X}_{pri}$, public dataset $\mathcal{X}_{pub}$, number of ensemble $N$, query $Q$, privacy parameter $\varepsilon$
\State Subsample and partition each dataset:
\State $\{D_i^{pri}\}_{i=1}^N \gets D^{pri}$, $\{D_i^{pub}\}_{i=1}^N \gets D^{pub}$
\For{$i= 1$  $\dots N$}
    \State Construct a few-shot prompt for each dataset:
    \State $P_{i}^s \gets \texttt{Prompt}(D_i^{s}, Q)$, \quad $s \in \{pri, pub\}$
    \State $O_i^s \gets \textbf{LLM}(P_{i}^s)$, \quad $s \in \{pri, pub\}$
\EndFor
\State Privately estimate cluster centres using DPM (Algorithm \ref{alg:DPM} of Appendix):
\State $\mathcal{C}, \text{weights} \gets \texttt{DPM}(\{O_i^{pri}\}_{i=1}^N \cup \{O_i^{pub}\}_{i=1}^N, \varepsilon)$
\State Sort cluster centers $\mathcal{C}$ by member count in decreasing order.
\State Choose the representative for each cluster: 
\State $S \gets \emptyset$
\For{$C \in \mathcal{C}$}
\State $i \gets \argmin_j \norm{C- O_j^{pub}}$
\State $S \gets S \cup \set{O_i^{pub}}$
\EndFor
\State Generate answer using top-$k$ candidates with a public 1-shot example:
\State $\mathbf{a} \gets \textbf{LLM}(\set{S_j}_{j=1}^k, D_{pub})$ \\ 
\Return $\mathbf{a}$
\end{algorithmic}
\end{algorithm}

\paragraph{Step 3: Final Response selection \& Generation} A straightforward approach to response generation is to select the representative of the cluster with the highest count. However, under strong privacy protection, the cluster with the second-highest count may incorrectly surpass the true highest-count cluster, potentially degrading model performance.
To obtain more reliable responses while leveraging public data as guidance, we pass the class representatives of the top-$k$ clusters to the selection stage. At this stage, we construct the prompt as:
\texttt{Select the most correct answer for the question from <candidates>},
with a public one-shot example prepended to the prompt (template in Appendix \ref{template}), as illustrated in Figure \ref{fig:model_overview}.

\subsection{Privacy Analysis}
In this section, we give analysis for calculating \texttt{DPM} parameter for the algorithm \ref{algo:privateICL} to be $(\varepsilon, \delta)$-DP. 
Assuming that the private clustering \texttt{DPM} Algorithm is $(\varepsilon, \delta)$-differentially private, Algorithm \ref{algo:privateICL} is also $(\varepsilon, \delta)$-differentially private. This follows from the post-processing property (definition \ref{postprocessing}), as lines 13–18 in the algorithm do not use any additional private data.  \\ 
In addition, since the model releases privatized output for each query, the privacy risk accumulates, which necessitates an accurate privacy loss tracking mechanism. In Algorithm \ref{algo:privateICL}, the private algorithm DPM employs \textit{exponential and Gaussian mechanisms} for private clustering. 

To trace the accurate privacy loss for the exponential mechanism, we express the privacy loss as Rényi Differential Privacy (RDP) using \citet{Bun2016ConcentratedDP}, then compose the privacy guarantees under RDP, and finally convert the result back to $(\varepsilon, \delta)$-DP using the theorem from \citet{Balle2019HypothesisTI}.  
For the Gaussian mechanism in DPM, we use \texttt{DPSGDAccount} from the \texttt{prv\_accountant} library, ensuring tight privacy tracking.

%% file: Sections/experiments.tex
\section{Experiments and Results}
In this section, we present the privacy-utility tradeoff of our private ICL models in question answering and summarization tasks. We begin by evaluating model performance using In-Distribution (ID) public data and Out-Of-Distribution (OOD) public data. Finally, we assess the empirical effectiveness of private ICL through a membership inference experiment. In addition to OOD public data, we also evaluate the performance of private ICL with noisy public data in the Appendix \ref{sec:noisyPublic}. 
\subsection{Experiment Setting}

\paragraph{Task} 
We evaluate the performance of the private ICL model on question-answering and dialogue summarization tasks. 
We conduct the ICL task with 100 test queries using a 4-shot and 100-ensemble setting, where the ensemble method is applied along with 100 private and 100 public examples. For the question-answering task, we use the OpenAI GPT-3.5-turbo model and the Davinci-002 model model for summarization.  

\paragraph{Dataset}
For the question-answering task, we use the ChatDoctor benchmark \citep{li2023chatdoctor}, which consists of questions and answers collected from dialogues between patients and doctors on icliniq.com. We sample 3,900 data examples from the dataset, using 2,600 examples as the private dataset and 1,300 examples as the public dataset. For the OOD public dataset, we use the HealthCareMagic subset of ChatDoctor which comes from a different platform.  

For the dialogue summarization task, we use the SAMSum dialogue summarization dataset \cite{gliwa-etal-2019-samsum}. From the training dataset, we also use 2,600 examples as the private demonstration dataset and 1,300 examples as the public demonstration dataset. For the OOD public dataset, we use dialogsum benchmark \citep{chen-etal-2021-dialogsum}. 
\begin{table*}[t]
\centering
\begin{subtable}{.8\textwidth}
    \centering
    \resizebox{\textwidth}{!}{%
    \begin{tabular}{c|c c c c c|c c}
    \toprule 
\textbf{Method} & \textbf{Metrics}  & \boldmath$\varepsilon=1$ & \boldmath$\varepsilon=3$ & \boldmath$\varepsilon=8$ & \boldmath$\varepsilon=\infty$ (Agg) & \boldmath$\varepsilon=0$  & \boldmath$\varepsilon=\infty$ (4-shot) \\ 
\midrule 
\multirow{4}{*}{SGA (top-k)} 
 & BLEU $\uparrow$ & \cellcolor{orange!30}\textbf{22.21\textsubscript{0.32}} & \cellcolor{orange!30}\textbf{23.37\textsubscript{0.18}} & \cellcolor{orange!30}\textbf{23.67\textsubscript{0.09}} & \cellcolor{orange!30}\textbf{26.01} & 21.61 & 23.43 \\ 
 & METEOR $\uparrow$ & \cellcolor{blue!30}\textbf{16.65\textsubscript{0.34}} & \cellcolor{blue!30}\textbf{17.13\textsubscript{0.21}} & \cellcolor{blue!30}\textbf{18.24\textsubscript{0.28}} & \cellcolor{blue!30}\textbf{20.15} & 17.01 & 18.81 \\ 
 & ROUGE-1 $\uparrow$ & \cellcolor{cyan!30}\textbf{25.91\textsubscript{0.24}} & \cellcolor{cyan!30}\textbf{25.75\textsubscript{0.33}} & \cellcolor{cyan!30}\textbf{27.01\textsubscript{0.14}} & \cellcolor{cyan!30}\textbf{30.67} & 28.28 & 28.98 \\
 & LLM-Judge $\uparrow$ & \cellcolor{violet!30}\textbf{2.82}\textsubscript{0.06} & \cellcolor{violet!30}\textbf{3.08}\textsubscript{0.04} & \cellcolor{violet!30}\textbf{3.11}\textsubscript{0.04} & \cellcolor{violet!30}\textbf{3.24}& 2.75 & 2.86 \\
\midrule
\multirow{4}{*}{SGA (top-1)} 
 & BLEU $\uparrow$ & 19.57\textsubscript{0.24} & 21.87\textsubscript{0.14} & 21.24\textsubscript{0.21} & 23.77 & 21.61 & 23.43 \\ 
 & METEOR $\uparrow$ & 14.79\textsubscript{0.29} & 15.87\textsubscript{0.17} & 15.62\textsubscript{0.23} & 16.85 & 17.01 & 18.81 \\ 
 & ROUGE-1 $\uparrow$ & 23.12\textsubscript{0.34} & 24.73\textsubscript{0.16} & 24.22\textsubscript{0.31} & 26.44 & 28.28 & 28.98 \\ 
 & LLM-Judge $\uparrow$ & 2.71\textsubscript{0.04} & 2.88\textsubscript{0.05} & 2.91\textsubscript{0.04} & 3.22 & 2.75 & 2.86 \\
\midrule 
\multirow{4}{*}{KSA} 
 & BLEU $\uparrow$ & 15.98\textsubscript{0.56} & 16.53\textsubscript{0.20} & 17.41\textsubscript{0.26} & 24.89 & 21.61 & 23.43 \\ 
 & METEOR $\uparrow$ & 13.38\textsubscript{0.43} & 13.35\textsubscript{0.09} & 14.11\textsubscript{0.22} & 18.80 & 17.01 & 18.81 \\ 
 & ROUGE-1 $\uparrow$ & 19.05\textsubscript{0.44} & 19.25\textsubscript{0.22} & 20.93\textsubscript{0.15} & 29.16 & 28.28 & 28.98 \\ 
 & LLM-Judge $\uparrow$ & 2.58\textsubscript{0.05} & 2.61\textsubscript{0.05} & 2.71\textsubscript{0.04} & 3.12 & 2.75 & 2.86 \\
\midrule
\multirow{4}{*}{KSA w/o public}
 & BLEU $\uparrow$ & 15.65\textsubscript{0.45} & 16.36\textsubscript{0.21} & 16.13\textsubscript{0.52} & 24.03 & 21.61 & 23.43 \\ 
 & METEOR $\uparrow$ & 12.80\textsubscript{0.36} & 13.35\textsubscript{0.10} & 13.23\textsubscript{0.41} & 18.15 & 17.01 & 18.81 \\ 
 & ROUGE-1 $\uparrow$ & 18.23\textsubscript{0.39} & 19.61\textsubscript{0.32} & 20.93\textsubscript{0.15} & 27.38 & 28.28 & 28.98 \\ 
 & LLM-Judge $\uparrow$ & 2.52\textsubscript{0.06} & 2.59\textsubscript{0.05} & 2.68\textsubscript{0.06} & 3.08 & 2.75 & 2.86 \\
 \bottomrule
    \end{tabular}
    }
    \caption{\textbf{Private ICL results in the QA task.} The best results are highlighted: \textcolor{orange!70}{BLEU}, \textcolor{blue!70}{METEOR}, \textcolor{cyan!70}{ROUGE-1}, \textcolor{violet!70}{LLM-Judge (1-5 scale)}.}
    \label{tab:qa}
\end{subtable}

\vspace{0.4cm} 

\begin{subtable}{.8\textwidth}
    \centering
    \resizebox{\textwidth}{!}{%
    \begin{tabular}{c|c c c c c|c c}
    \toprule 
\textbf{Method} & \textbf{Metrics}  & \boldmath$\varepsilon=1$ & \boldmath$\varepsilon=3$ & \boldmath$\varepsilon=8$ & \boldmath$\varepsilon=\infty$ (Agg) & \boldmath$\varepsilon=0$  & \boldmath$\varepsilon=\infty$ (4-shot) \\ 
\midrule 
\multirow{4}{*}{SGA (top-k)} 
 & ROUGE-1 $\uparrow$ & \cellcolor{yellow!30}\textbf{36.58\textsubscript{0.59}} & 38.84\textsubscript{0.23} & \cellcolor{yellow!30}\textbf{39.90\textsubscript{0.71}} & 41.39 & 32.47 & 37.24 \\ 
 & ROUGE-2 $\uparrow$ & 13.34\textsubscript{0.50} & 14.03\textsubscript{0.54} & 15.93\textsubscript{0.21} & 16.49 & 10.72 & 12.86 \\ 
 & ROUGE-L $\uparrow$ & \cellcolor{cyan!30}\textbf{29.36\textsubscript{0.83}} & 29.17\textsubscript{0.69} & \cellcolor{cyan!30}\textbf{31.62\textsubscript{0.12}} & \cellcolor{cyan!30}\textbf{32.91} & 25.50 & 28.85 \\ 
 & LLM-Judge $\uparrow$ & \cellcolor{violet!30}\textbf{3.25}\textsubscript{0.04} & 3.23\textsubscript{0.05} & \cellcolor{violet!30}\textbf{3.33}\textsubscript{0.03} & \cellcolor{violet!30}\textbf{3.37} & 3.01 & 3.18 \\
\midrule
\multirow{4}{*}{SGA (top-1)} 
 & ROUGE-1 $\uparrow$ & 36.48\textsubscript{0.56} & 37.88\textsubscript{0.57} & 37.32\textsubscript{0.75} & \cellcolor{yellow!30}\textbf{41.59} & 32.47 & 37.24 \\ 
 & ROUGE-2 $\uparrow$ & 12.18\textsubscript{0.32} & 13.34\textsubscript{0.21} & 13.02\textsubscript{0.64} & 16.26 & 10.72 & 12.86 \\ 
 & ROUGE-L $\uparrow$ & 27.93\textsubscript{0.47} & 29.46\textsubscript{0.33} & 28.57\textsubscript{0.31} & 32.10 & 25.50 & 28.85 \\ 
 & LLM-Judge $\uparrow$ & 3.15\textsubscript{0.05} & 3.17\textsubscript{0.08} & 3.22\textsubscript{0.06} & 3.35 & 3.01 & 3.18 \\
\midrule 
\multirow{4}{*}{KSA} 
 & ROUGE-1 $\uparrow$ & 35.06\textsubscript{0.73} & \cellcolor{yellow!30}\textbf{39.86}\textsubscript{0.59} & 39.63\textsubscript{0.34} & 41.10 & 32.47 & 37.24 \\ 
 & ROUGE-2 $\uparrow$ & \cellcolor{green!30}\textbf{14.52\textsubscript{0.31}} & \cellcolor{green!30}\textbf{16.95\textsubscript{0.48}} & \cellcolor{green!30}\textbf{16.63\textsubscript{0.41}} & \cellcolor{green!30}\textbf{18.06} & 10.72 & 12.86 \\ 
 & ROUGE-L $\uparrow$ & 27.48\textsubscript{0.25} & \cellcolor{cyan!30}\textbf{32.08\textsubscript{0.67}} & 31.41\textsubscript{0.35} & 32.47 & 25.50 & 28.85 \\ 
 & LLM-Judge $\uparrow$ & 3.23\textsubscript{0.04} & \cellcolor{violet!30}{3.28\textsubscript{0.03}} & 3.28\textsubscript{0.04} & 3.36 & 3.01 & 3.18 \\
\midrule
\multirow{4}{*}{KSA w/o public}
 & ROUGE-1 $\uparrow$ & 33.45\textsubscript{0.88} & 35.90\textsubscript{0.59} & 38.10\textsubscript{0.54} & 39.24 & 32.47 & 37.24 \\ 
 & ROUGE-2 $\uparrow$ & 12.88\textsubscript{0.78} & 14.78\textsubscript{0.48} & 15.71\textsubscript{0.58} & 16.45 & 10.72 & 12.86 \\ 
 & ROUGE-L $\uparrow$ & 26.64\textsubscript{0.63} & 28.69\textsubscript{0.54} & 30.59\textsubscript{0.62} & 31.21 & 25.50 & 28.85 \\ 
 & LLM-Judge $\uparrow$ & 3.15\textsubscript{0.04} & 3.17\textsubscript{0.03} & 3.22\textsubscript{0.04} & 3.35 & 3.01 & 3.18 \\
 \bottomrule
    \end{tabular}
    }
    \caption{\textbf{Private ICL results in the summarization task.} The best results are highlighted: \textcolor{yellow!75!orange}{ROUGE-1}, \textcolor{green!70}{ROUGE-2}, \textcolor{cyan!70}{ROUGE-L}, , \textcolor{violet!70}{LLM-Judge (1-5 scale)}.}
    \label{tab:summarization}
\end{subtable}
\caption{Private ICL methods run for each $\varepsilon=1,3,8$. $\varepsilon=\infty$ denotes ensemble non-private methods, \textbf{$\varepsilon=0$ denotes $4$-shot with OOD public}, a\textbf{nd $\varepsilon=\infty$ denotes $4$-shot prediction with private demonstration examples}. Results for DP algorithms are averaged over $5$ runs with different seeds. SGA top-$1$ denotes the top-$1$ selection without the candidate selection. KSA w/o public denotes the KSA method with only private data.}
\vspace{-0.1cm}
\end{table*}
\paragraph{Models}
We evaluate four models, including baselines: \textbf{SGA (top-$k$)}, our private ICL model that applies private clustering in the semantic embedding space, where embeddings are generated by the \texttt{text-embedding-ada-002} model; \textbf{SGA (top-1)}, which also performs private clustering in the semantic space but directly selects the top-1 element without candidate selection; \textbf{KSA}, a baseline private ICL model with keyword space aggregation \citep{wu2024privacypreserving}; and \textbf{KSA w/o public}, which performs keyword space aggregation using only private data.

We evaluate the models at three privacy levels: $\varepsilon=1, 3, 8$, corresponding to strong, moderate, and weak privacy protection, respectively. We also include two \red{non-private} baselines and one \blue{fully private} baseline: a 4-shot model with sampled private examples (\red{$\varepsilon = \infty$}), a 4-shot model using non-private aggregation (\red{$\varepsilon = \infty$ (Agg)}) and a 4-shot model using OOD public (\blue{$\varepsilon=0$}). 
The hyperparameter settings for privacy analysis are summarized in Table \ref{tab:hyperparameter} of the Appendix. 

\begin{figure*}[t]
    \centering

    \begin{subfigure}{.85\textwidth}
        \centering
        \includegraphics[width=\linewidth]{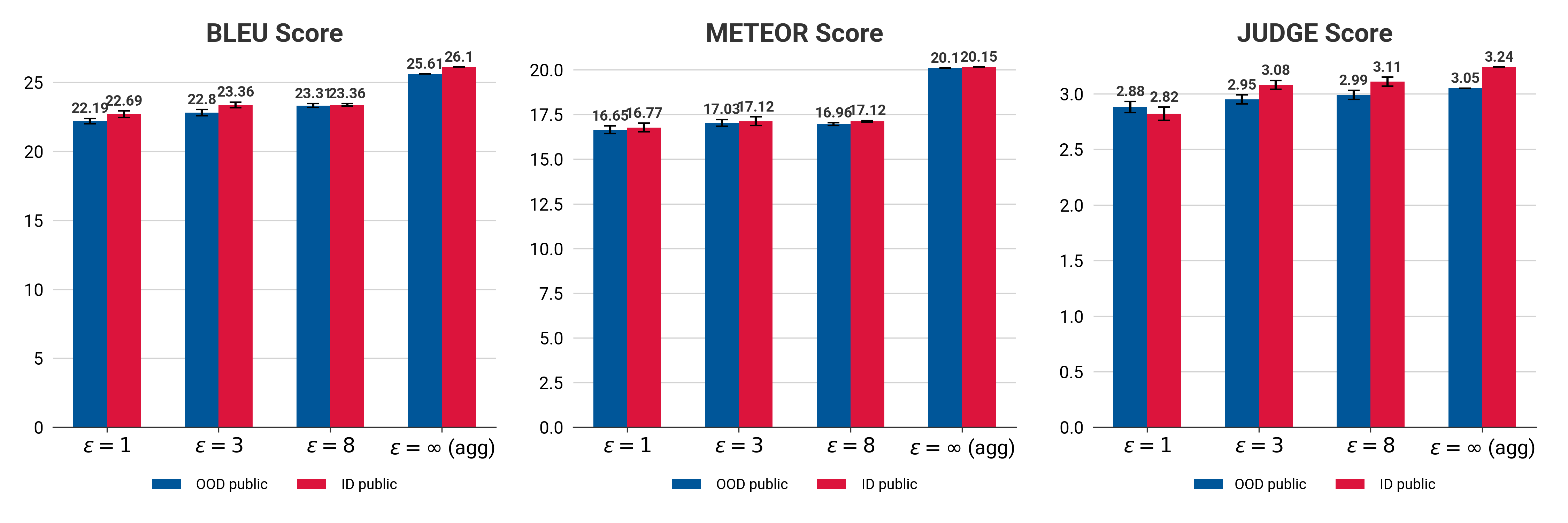}
        \caption{Question answering task}
        \label{fig:subfig1}
    \end{subfigure}
    \vspace{-0.2cm}
    \begin{subfigure}{.85\textwidth}
        \centering
        \includegraphics[width=\linewidth]{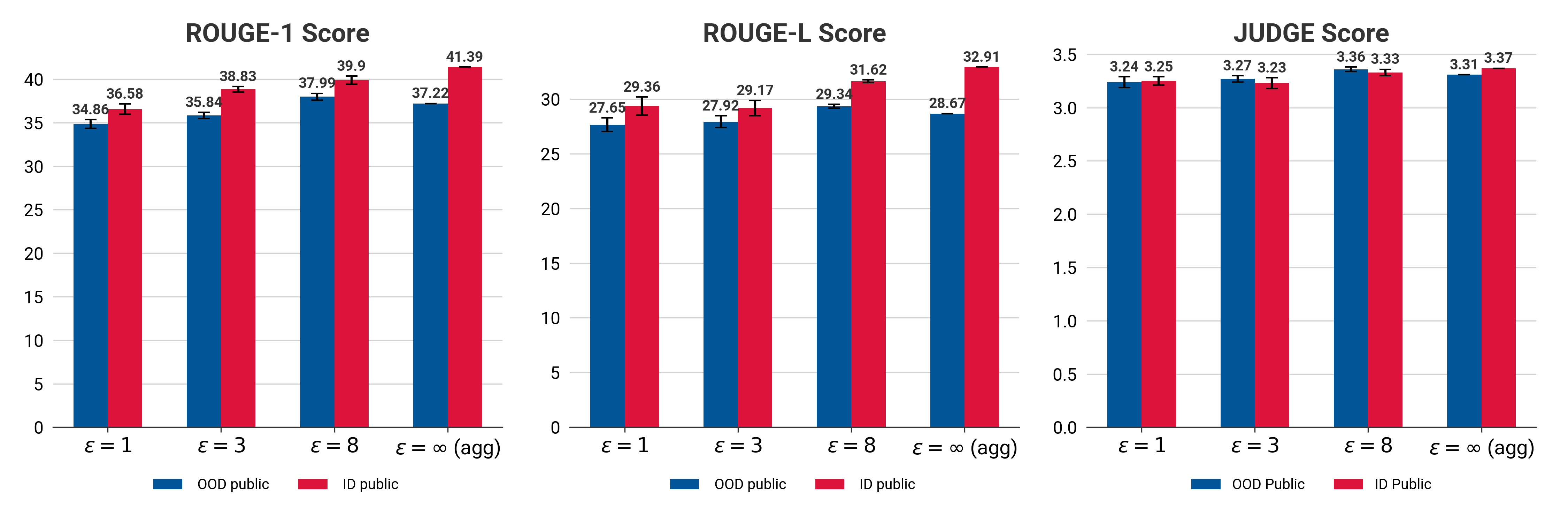}
        \caption{Summarization task}
        \label{fig:subfig2}
    \end{subfigure}
    \caption{\textbf{Results for SGA model with OOD public dataset} OOD public refers to
private ICL with out-of-distribution public data and ID public to private ICL with in-distribution public data.}
    \label{fig:ood}
\end{figure*}
\subsection{Private ICL with ID public}
\paragraph{Question Answering Task}
The results for the question answering task are presented in Table \ref{tab:qa}, evaluated using BLEU, METEOR, ROUGE-1 and LLM-Judge metrics. For the LLM-Judge evaluations, we employ GPT-4o as a judge to assess the generation quality of each model on the QA task.  From the experiment, we observe the following: 
(1) \textbf{Candidate selection guided by public examples leads to better results}. For all privacy levels $\varepsilon$, SGA with top-$k$ selection outperforms the SGA top-$1$ method, demonstrating the usefulness of public data. (2) \textbf{The SGA method outperforms the KSA method across all privacy levels.} We assume that the KSA method struggles with reconstructing entire answers with noisy keywords, as the answer length in the ChatDoctor benchmark is typically more than 3 sentences. (3) \textbf{Private aggregation with public data improves performance.} The KSA method with public data outperforms the KSA method without public data across all privacy budgets $\varepsilon$. Finally, we observe that $\varepsilon=\infty$ with aggregation outperforms the direct $4$-shot method.

\paragraph{Summarization Task}
The summarization results are presented in Table \ref{tab:summarization} with ROUGE-1,2,L and LLM-Judge metrics.
From the results, we observe the following: (1)\textbf{ Both SGA and KSA show reasonable performance with a strong privacy budget ($\varepsilon=1$)}, with comparable performance to 4-shot  $\varepsilon=\infty$ without aggregate. (2) \textbf{Using public data for candidate selection and aggregation improves performance.} The SGA method with top-$k$ selection outperforms the top-1 SGA, and the KSA method outperforms KSA without public data. 
(3) \textbf{SGA shows more robust performance at a strong privacy regime $(\varepsilon=1)$.} The robustness of SGA may be attributed to the fact that KSA reconstructs the summary from noisy keywords, whereas SGA generates a summary from previously generated summaries.
\vspace{-0.2cm}
\subsection{Private ICL with OOD public}
We utilize the best-performing model, SGA, to evaluate private ICL using an OOD public dataset, with the results for question answering and summarization tasks shown in Figure \ref{fig:ood}. Although ID public data consistently achieves slightly higher scores across most metrics, the OOD public data also demonstrates strong and competitive performance, particularly noteworthy in the question-answering task, where the differences are minimal.

In the summarization task, while the performance gap is more pronounced at higher privacy budgets, OOD data performance notably improves as the privacy budget tightens. This observation indicates that OOD public data remains highly valuable and delivers robust performance, particularly under stricter privacy constraints. Thus, even ICL with OOD data, the SGA method maintains commendable effectiveness, underscoring the practical utility of OOD datasets in privacy-sensitive scenarios.

\subsection{Membership Inference Attack}
For the empirical evaluation of privacy protection,  we adopt a widely used membership inference attack (MIA) against the ICL framework \cite{Shokri2016MembershipIA}. The attacker's goal is to determine whether a target example is part of the demonstration examples used for ICL. We implement the repeat attack method \cite{Wen2024MembershipIA}. The intuition behind this attack is that the model tends to complete the training sentence when provided with only the first few words of the target example. The inference attack procedure works as follows: (1) The attacker \textbf{selects a target example} and attempts to determine whether it is part of the training dataset. (2) The attacker \textbf{truncates the target example} and inputs it into the model, which generates a completion. (3) The attacker \textbf{calculates the semantic similarity} between the model’s completion and the target example. If the similarity exceeds a threshold $\tau$, the attacker concludes that the target example was part of the training set.

\begin{figure}[h]
\centering \includegraphics[width=0.9\linewidth]{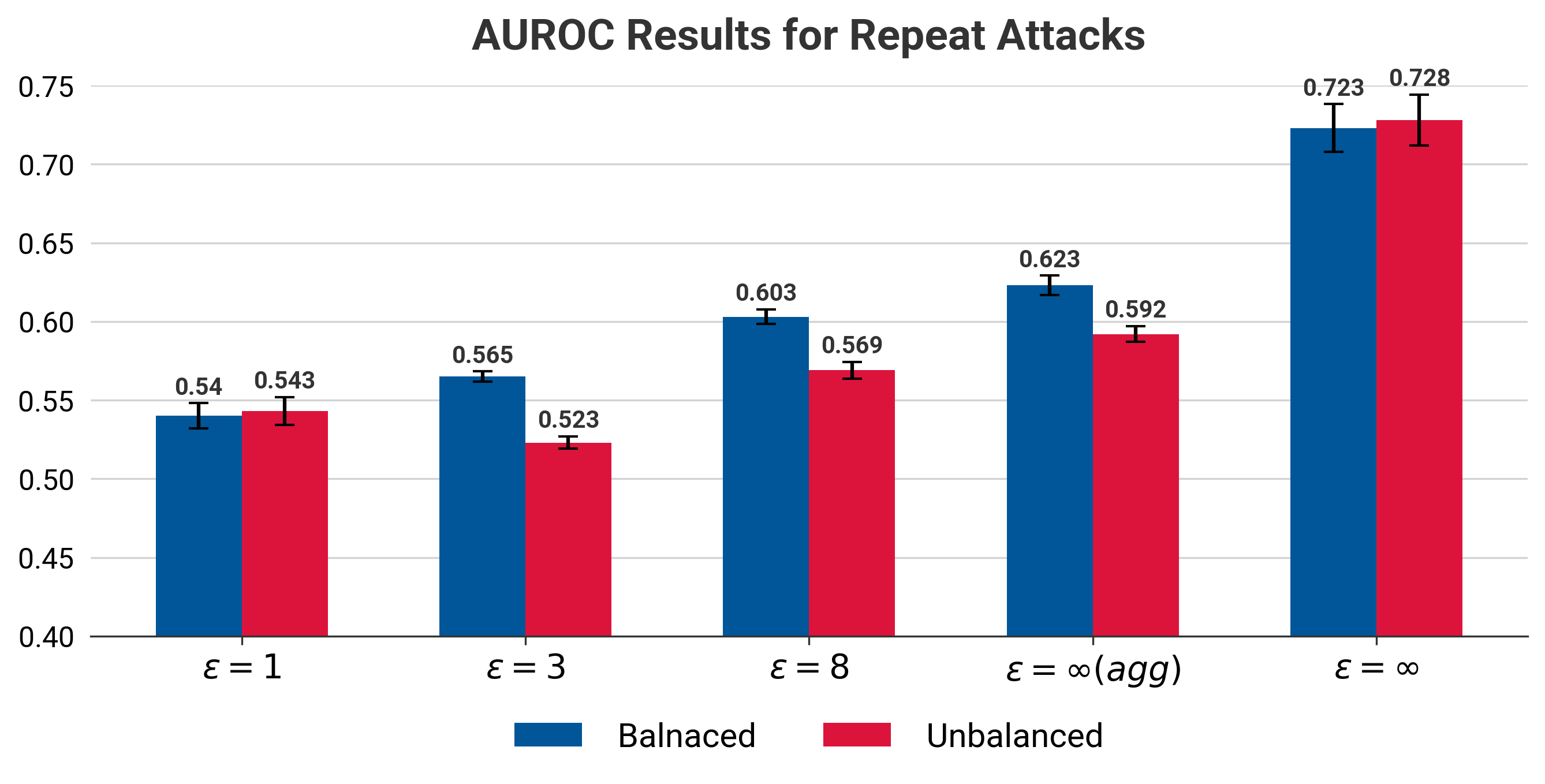} \caption{AUROC results for repeat attacks. "Balanced" denotes the MIA setting where the ratio between members and non-members is $1:1$, and "Unbalanced" denotes a ratio of $1:4$.} \label{fig:attack} 
\end{figure}

We experiment with two different member-to-non-member ratios: (1) \textbf{Balanced scenario}: The number of member and non-member examples is equal, with a ratio of $1:1$. (2) \textbf{Unbalanced scenario}: The ratio is $1:4$, reflecting a more realistic environment. For the balanced setting, we use 40 member examples and 40 non-member examples. For the unbalanced setting, we use 40-member examples and 160 non-member examples. We use a 2-shot, 10-ensemble setting for the private model and apply SGA with $\varepsilon = [1, 3, 8]$. For the non-private baselines, we consider the non-private aggregation model ($\varepsilon=\infty$, agg) and a 20-shot model ($\varepsilon=\infty$), since the effective number of shots for the private model is $2 \times 10 = 20$. The AUROC is computed over member and non-member examples, with the results shown in Figure \ref{fig:attack}.

From the results, we observe that, at all $\varepsilon$ levels, the private models are robust to the repeat attack, keeping the AUROC around or below $0.6$, indicating low attack success. Notably, the non-private aggregation model also provides some defense, achieving an AUROC of $0.592$ for the unbalanced setting. This suggests that non-private aggregation introduces a degree of empirical privacy, possibly due to the aggregation process reducing the influence of individual examples, even though it lacks formal differential privacy (DP) guarantees.

\subsection{Private ICL with noisy public data}
\label{sec:noisyPublic}
We conduct an experiment on private ICL using misaligned public data. Specifically, we collected public examples that are furthest from the privatized centroids, where the corresponding ICL examples are non-informative, and then instructed the GPT model to generate additional data based on these noisy public examples. This setup enables us to systematically evaluate the robustness of our method under degraded public data quality.
The following table shows BLEU scores under varying amounts of noisy public data:
\begin{table}[h]
\centering
\resizebox{\linewidth}{!}{%
\begin{tabular}{lcccc}
\toprule
\textbf{Data Type} & \multicolumn{4}{c}{\textbf{BLEU Scores under Privacy Budget}} \\
\cmidrule{2-5}
& $\epsilon=1$ & $\epsilon=2$ & $\epsilon=4$ & $\epsilon=8$ \\
\midrule
100 noisy public   & 17.43 & 19.88 & 20.89 & 21.98 \\
500 noisy public   & 19.26 & 20.79 & 21.12 & 22.68 \\
2000 noisy public  & 20.79 & 21.14 & 21.51 & 22.41 \\
ID Public          & 22.21 & 23.36 & 23.67 & 26.01 \\
\bottomrule
\end{tabular}
}
\caption{Performance evaluation with varying amounts of noisy public data.}
\label{tab:noisy_public_bleu}
\end{table}
While augmentation with more noisy public data does not help under a high privacy budget, it still improves performance under a low privacy budget (e.g., $\epsilon=1$ or $\epsilon=2$).

%% file: Sections/Analysis.tex
\begin{figure}[h]
    \centering
    \includegraphics[width=0.9\linewidth]{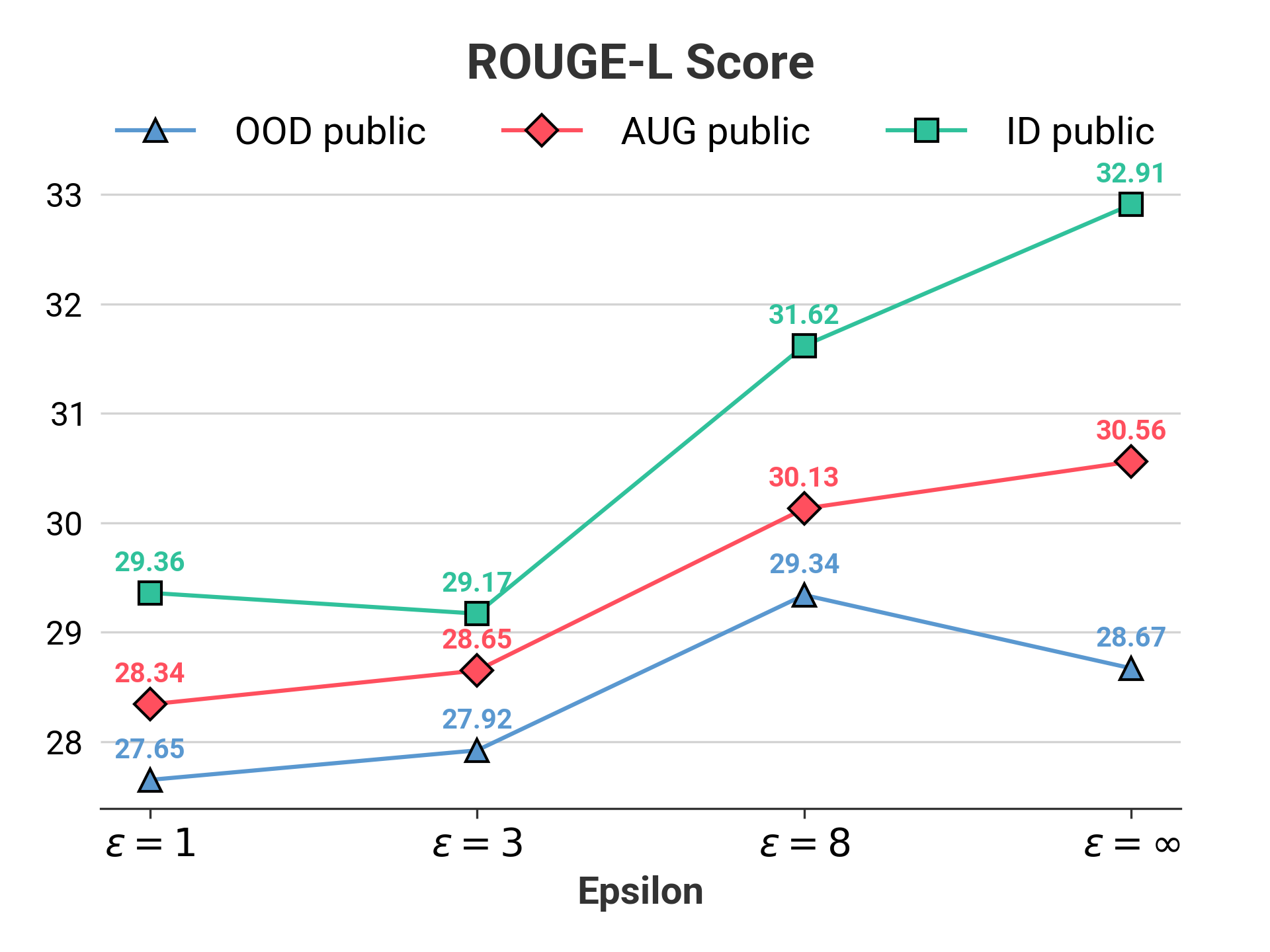}
    \caption{ROUGE-L score with augmented public dataset referred as "AUG public" in the summarization task.}
    \label{fig:aug}
\end{figure}

\section{Enhancing Utility and Efficiency in Private ICL}
\subsection{Public Data Quality Enhancement}
In many domains, high-quality public datasets may be unavailable or inferior to the private datasets at hand, making it challenging to apply our framework. To address this issue, we allocate a small privacy budget to augment the public dataset. Our augmentation follows a simple approach: first, we collect public examples that are closest to privatized centroids with a budget of $\varepsilon=1$. Then, we instruct the GPT-Turbo model to generate additional data based on these filtered public examples. 

The performance of the proposed augmentation technique on the summarization task is reported in Figure \ref{fig:aug}. The results show that even with a small privacy budget, the augmentation method is effective, showing about $1\sim2$ ROUGE-L score difference compared to ID public setting and outperforming the baseline that relies solely on OOD public datasets. Finally, when no public data is available, a private generator can be an effective solution by synthesizing samples for augmentation.
\subsection{Accelerating Private ICL with Coreset Sampling}
\label{sec:FA}

\begin{figure}
    \centering
    \includegraphics[width=1.0\linewidth]{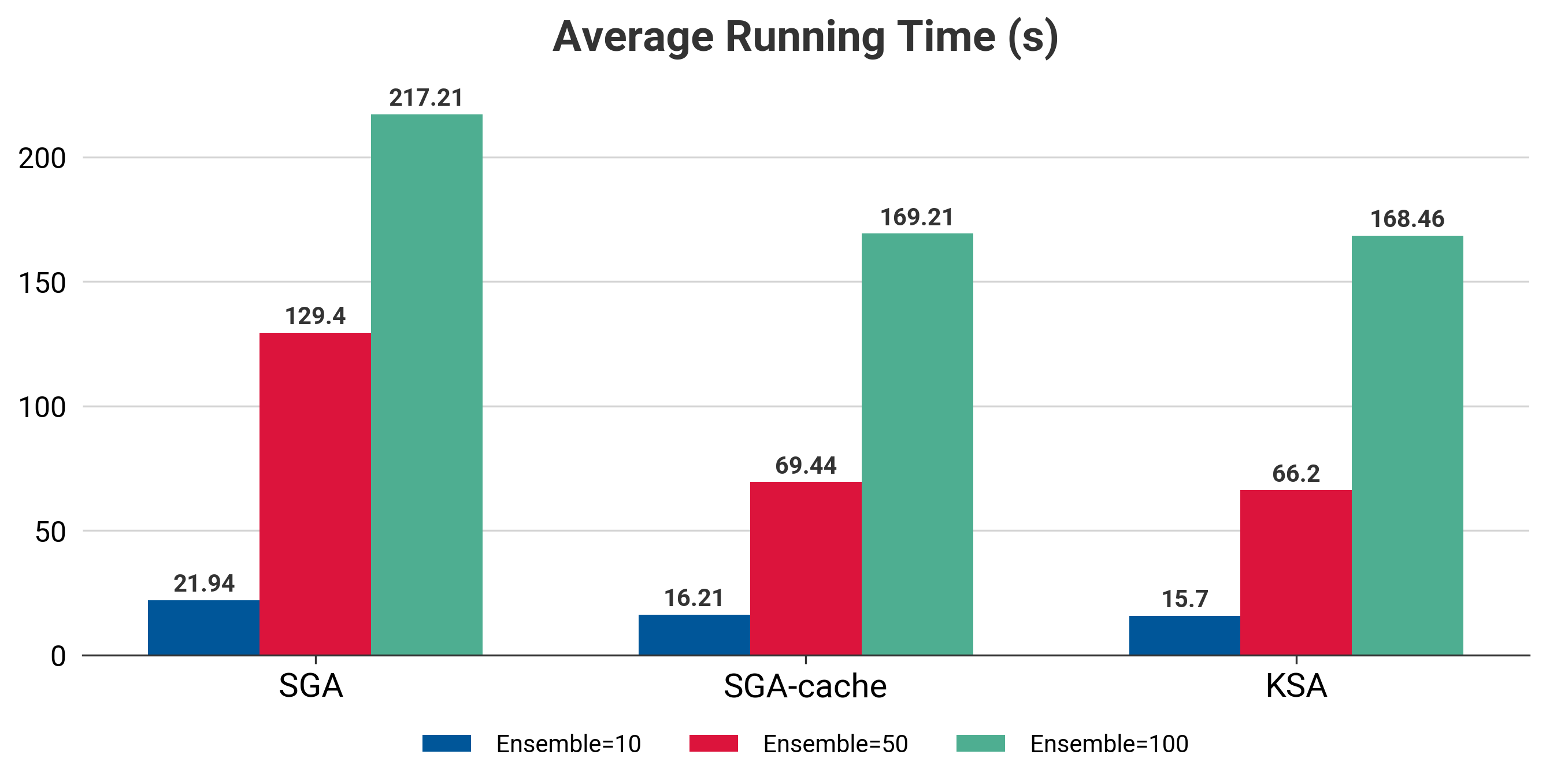}
    \caption{Average running time of SGA and KSA. SGA with cache denotes the SGA with precomputed embeddings.}
    \label{fig:comp}
\end{figure}
\begin{figure}
    \centering
    \includegraphics[width=1.0\linewidth]{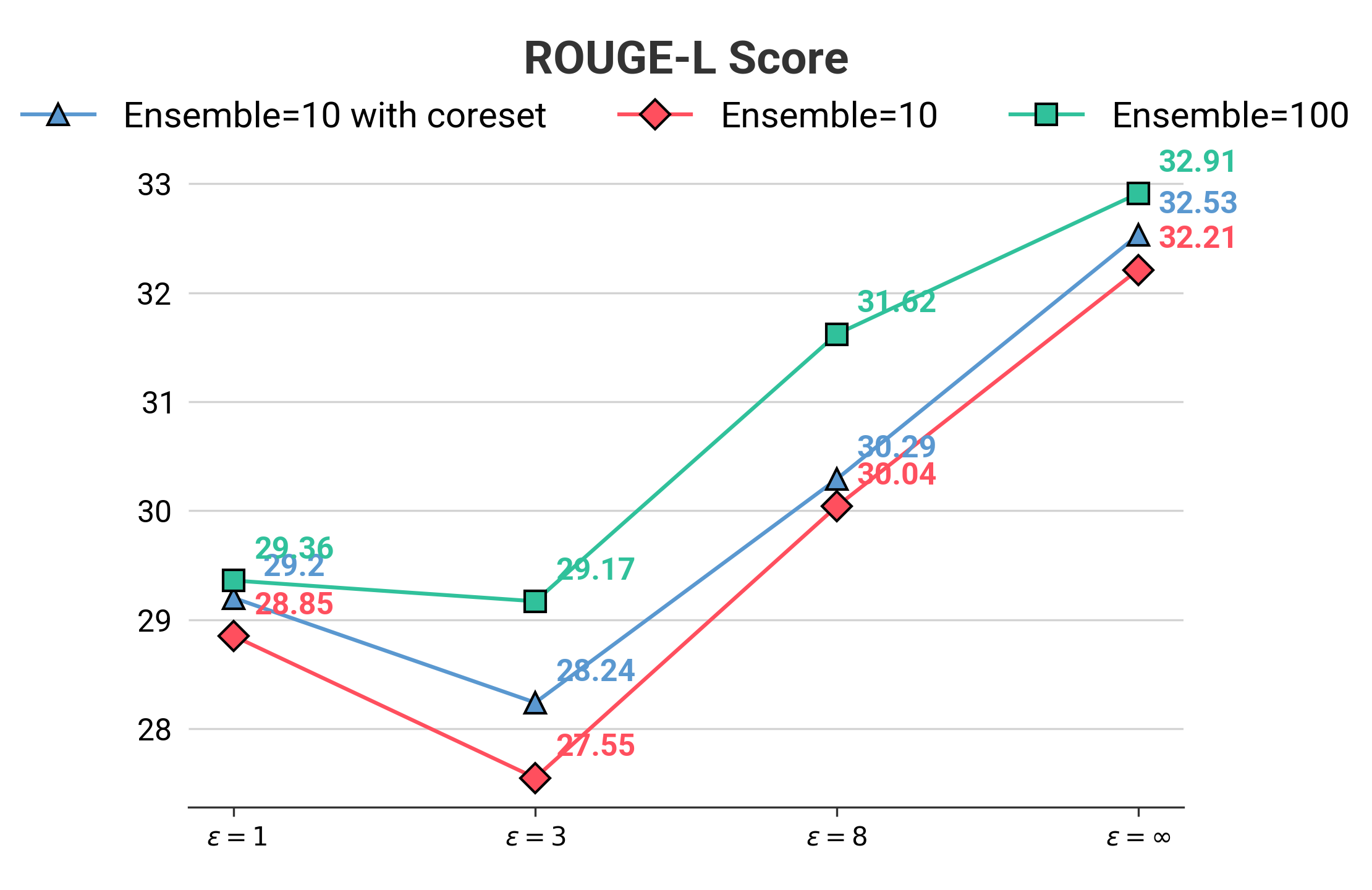}
    \caption{ROUGE-L scores across different privacy budgets ($\epsilon$) for three methods: Ensemble=10 with coreset, Ensemble=10, and Ensemble=100.}
    \label{fig:coreset}
    \vspace{-0.4cm}
\end{figure}
The computation time for private ensembles scales with the number of ensembles (Figure \ref{fig:comp}). Results indicate that the SGA method with pre-embedding computation (SGA-cache) significantly improves computation efficiency. To further reduce computational complexity, we employ a simple technique called \emph{coreset-sampling}. Specifically, we cluster the ICL samples into distinct groups using K-Means clustering based on text embeddings, selecting the centroids of each cluster as a representative subsampled coreset. This coreset effectively captures the characteristics of the entire batch ensemble.
Performance with the coreset approach is shown in Figure \ref{fig:coreset}. We observe that ensemble size 10 with coreset sampling achieves improved performance compared to ensemble size 10 with random sampling. 
\vspace{-0.4cm}
\begin{figure}[ht]
\end{figure}

%% file: Sections/conclusion.tex
\section{Conclusion}
We propose a private in-context learning (ICL) framework that leverages public data. To address high dimensionality, we project the LLM output into a semantic group space. Public data is used in ensemble aggregation and selection to mitigate utility degradation from differential privacy. Experiments show an effective privacy-utility tradeoff, with notable improvements from public data. Additionally, empirical tests demonstrate strong defense against privacy and membership inference attacks across all privacy levels. Finally, we present techniques for public data enhancement and inference acceleration, offering a practical solution for real-world applications.

%% file: Sections/limitation.tex
\section*{Limitations}
One limitation of the private ICL framework is that the privacy risk accumulates over multiple queries. While we attempt to mitigate utility degradation caused by accumulated query outputs by leveraging public data, this approach may not entirely prevent privacy leakage in long-running or high-volume query settings. We also believe that this privacy accumulation issue could be alleviated by updating private demonstration examples once a query budget threshold is reached.

Another limitation is that the private ICL framework requires extensive computation due to the use of multiple ensembles. As we have demonstrated, techniques such as coreset sampling can be employed to reduce the computational burden. Future work could explore more computationally efficient differential privacy mechanisms or optimized ensemble methods to further alleviate computational costs without compromising privacy guarantees.

Lastly, public data may not always be accessible in certain domains, especially in sensitive areas where such data is limited or unavailable. As we have demonstrated, this issue can be partially alleviated by synthesizing new examples while investing only a small privacy budget

\section*{Ethical Considerations}

Data privacy is a critical ethical concern in in-context learning frameworks, especially those leveraging private demonstration examples. Our proposed approach incorporates differential privacy to ensure robust privacy guarantees, minimizing the risk of sensitive information leakage from demonstration data. Throughout our experiments, we exclusively utilized publicly available, open-source doctor-patient dialogues, thereby avoiding potential ethical issues related to private data acquisition and usage.

Despite our strict adherence to public data in experimental settings, real-world deployments of similar frameworks may involve sensitive personal information. We emphasize the necessity of carefully applying differential privacy mechanisms to balance utility and privacy, ensuring compliance with ethical guidelines and privacy regulations such as GDPR and HIPAA.

\section*{Acknowledgments}
This work was partly supported by the Institute of Information \& Communications Technology Planning \& Evaluation (IITP) grant funded by the Korea government (MSIT) [No. RS-2023-00229780, Development of Artificial Intelligence Technology for Process-focused Evaluation (Student’s Learning Diagnosis); No. RS-2021-II211343, Artificial Intelligence Graduate School Program (Seoul National University); No. RS-2021-II212068, Artificial Intelligence Innovation Hub (Artificial Intelligence Institute, Seoul National University)]; and
the BK21 FOUR program of the Education and
Research Program for Future ICT Pioneers, Seoul
National University in 2024. K. Jung is with ASRI, Seoul National University, Korea.

%% file: Sections/appendix.tex
\newpage 
\appendix 
\section{Details of Differential Privacy}
\subsection{DPM algorithm}
For the private clustering algorithm, we use DPM from \citep{liebenow2024dpmclusteringsensitivedata}. DPM recursively splits a dataset into clusters by identifying sparse regions while preserving differential privacy (DP). We allocate the privacy budget for the Gaussian mechanism with $\varepsilon_{\text{avg}}$ and for the exponential mechanism with $\varepsilon_{\text{exp}}$. For ease of analysis, unlike the original paper, we assume that the size of the dataset and the size of the interval are \emph{public}. By excluding them from the privacy budget, we can focus on protecting more sensitive aspects of the data while simplifying the analysis. 
\subsection{Privacy Accounting}
As privacy accounting involves composition, we utilize the following theorems for tight accounting. 
 \begin{theorem}[\citet{Balle2018PrivacyAB}]
     Let \texttt{Uniform} denotes sampling $m$ elements from $n$ data points without replacement
     Let $\cM^{\prime} = \cM \circ \texttt{Uniform}$. For any $\varepsilon \geq 0 $ we have $\delta_{\cM^{\prime}} (\varepsilon^{\prime}) \leq (m/n) \delta_{\cM} (\varepsilon)$, where $\varepsilon^{\prime} = \log (1+ (m/n) (e^{\varepsilon} -1))$. 
\label{thm:amplification}
\end{theorem}

 \begin{theorem}[DP to RDP \citet{Bun2016ConcentratedDP}]
     The exponential mechanism is $\varepsilon$-DP and $(\alpha, \varepsilon_{\texttt{EM}}(\alpha))$-RDP, where $\varepsilon_{\texttt{EM}} (\alpha)$ is defined as
\begin{equation*}
  \resizebox{\linewidth}{!}{$
    \min \left( \frac{\alpha}{2} \varepsilon^2, \frac{1}{\alpha - 1} \log \left( \frac{\sinh (\alpha \varepsilon) - \sinh ((\alpha - 1) \varepsilon)}{\sinh (\varepsilon)} \right) \right)
  $}
\end{equation*}
\label{thm:DP2RDP}
\end{theorem}
\vspace{-1.5cm}
\begin{theorem}[RDP composition]
If each mechanism $(\cM_i)_{i=1}^k$ is $(\alpha, \varepsilon)$-RDP, then the composition $A_k \circ A_{k-1} \circ \dots \circ A_1$ is $(\alpha, k \varepsilon)$-RDP.  
\end{theorem}

\begin{theorem}[RDP to approximate DP  \citet{Balle2019HypothesisTI}]

If a mechanism $\cM$ is $(\alpha, \rho)$-RDP then it is $(\rho + \log ( (\alpha-1) / \alpha) - (\log \delta + \log \alpha) /  (\alpha-1), \delta)$-DP for any $0 < \delta <1$. 
\label{thm:RDP2DP}
\end{theorem}

When accounting exponential mechanism of DPM, we first apply theorem \ref{thm:DP2RDP}. Then the calculate privacy loss of composition by calling \texttt{compose\_subsampled\_EM} API using \texttt{AutoDP} package. Then, the converting RDP to approx DP using theorem \ref{thm:RDP2DP}. For the Gaussian mechanism of DPM, we use \texttt{DPSGDAccount} from the \texttt{prv\_accountant} library. 

\begin{algorithm}[H]
\caption{\texttt{DPM}, \citep{liebenow2024dpmclusteringsensitivedata}}
\label{alg:DPM}
\begin{algorithmic}[1]
\Require$D, \tau_r, R, t, q, \alpha, \varepsilon_{\text{int}}, \varepsilon_{\text{cnt}},  \varepsilon_{\text{exp}}, \varepsilon_{\text{avg}}$
\State clusters $\gets \emptyset$ 
\State weights $\gets \emptyset$
\State $\left( \varepsilon_{\text{cnt},i} \right)_{i=0}^{\tau_r} = \left( \varepsilon_{\text{cnt}} \frac{\sqrt{2^i}}{\sum_{j=0}^{\tau_r} \sqrt{2^j}} \right)_{i=0}^{\tau_r}$
\State $\left( \varepsilon_{\text{exp},i} \right)_{i=0}^{\tau_r-1} = \left( \varepsilon_{\text{exp}} \frac{\sqrt{2^i}}{\sum_{j=0}^{\tau_r-1} \sqrt{2^j}} \right)_{i=0}^{\tau_r-1}$
\State $\left( \lambda_i \right)_{i=0}^{\tau_r} = \left( -\ln(2\delta) / \varepsilon_{\text{cnt},i} \right)_{i=0}^{\tau_r}$ 
\LineComment{Dataset size perturbation}
\State $\tilde{n} = |D| + \text{Lap}(\varepsilon_{\text{cnt},0})$
\LineComment{Private interval size estimation}
\State $\beta = \text{IntervalSizeEst}(D, \tilde{n}, \varepsilon_{\text{int}}, \text{sigmas})$ 
\State $\text{numSplits} = (b - a) / \beta$ 
\State \Call{BuildClustering}{$D, \tilde{n}, 0$}
\LineComment{Privately compute the cluster \newline centers}
\State $C = \{ \text{DPAvg}(C_i, \tilde{n}, \varepsilon_{\text{avg}}) \mid C_i \in \text{clusters}\}$
\State \Return $C, \text{weights}$

\Procedure{BuildClustering}{$S, \tilde{n}, y$}
    \If {$y \geq \tau_r$} 
        \State halt and add $S$ to clusters and $\tilde{n}$ to weights
    \EndIf
    \State $S_1, S_2 = \text{Split}(S, \tilde{n}_S, y)$
    \State $\tilde{n}_{S_1} = |S_1| + \text{Lap}(\varepsilon_{\text{cnt},y+1})$
    \State $\tilde{n}_{S_2} = |S_2| + \text{Lap}(\varepsilon_{\text{cnt},y+1})$
    \If {$\tilde{n}_{S_i} < \tau_e$} 
        \State halt and add $S$ to clusters and $\tilde{n}$ to weights
    \EndIf
    \State \Call{BuildClustering}{$S_1, \tilde{n}_{S_1}, y+1$}
    \State \Call{BuildClustering}{$S_2, \tilde{n}_{S_2}, y+1$}
\EndProcedure

\Procedure{Split}{$S, \tilde{n}, y$}
    \State $\Delta_f = \frac{t/q + \alpha}{\tilde{n} - \lambda_y}$
    \LineComment{Private arg-max to find best split index $i^*$}
    \State $i^* = M_E(S, f, \varepsilon_{\text{exp},y})$ 
    \State $d^* = \lfloor d \cdot \text{numSplits} / i^* \rfloor$
    \State $s^* = \left( (d \cdot \text{numSplits} \mod i^*) + 0.5 \right) \cdot \beta$
    \State $S_1 = \{ x \in S \mid x^{(d^*)} \leq s^* \}$
    \State $S_2 = \{ x \in S \mid x^{(d^*)} > s^* \}$
    \State \Return $S_1, S_2$
\EndProcedure
\end{algorithmic}
\end{algorithm}

\section{Further Analysis}
\subsection{Candidate Number Ablation Study}
In this experiment, conducted on the ChatDoctor Benchmark, we evaluate the top-$k$ SGA model under varying candidate numbers $k$ and privacy budgets. The results are summarized in Table~\ref{tab:k_bleu}. At $\epsilon=8$, there are no noticeable differences across different $k$ values. However, for $k=2$ and $k=8$, the model’s performance degrades more rapidly in the low-$\epsilon$ regime compared to $k=3$ and $k=6$. We conjecture that a small number of candidates ($k=2$) does not provide sufficient options to mitigate the effects of noisy histogram outputs, while a large number of candidates ($k=8$) introduces excessive confusion for the model, leading to performance degradation.
\begin{table}[h]
\centering
\begin{tabular}{lccc}
\toprule
\textbf{BLEU} & $\varepsilon{=}1$ & $\varepsilon{=}3$ & $\varepsilon{=}8$ \\
\midrule
$k{=}2$ &  23.88 & 23.91  & 24.88   \\
$k{=}3$ &  23.98 & 24.39  & 24.52  \\
$k{=}6$ &  24.81 & 24.78  & 25.41  \\
$k{=}8$ &  24.12 & 24.65  & 25.38  \\
\bottomrule
\end{tabular}
\caption{Comparison of BLEU scores across candidate number $k$ and privacy budgets.}
\label{tab:k_bleu}
\end{table}

\subsection{Open Source Model}
In this experiment, we evaluate the performance of our private ICL framework against repeat attacks targeting LLaMA-8B-Instruct, one of the widely used open-source LLMs. Since our method is model-agnostic and assumes a black-box setting, the SGA model is applied without specific modification. The results are presented in Table~\ref{tab:auroc_repeat_llama}. Overall, the findings are consistent with the evaluation on the GPT series: the AUROC decreases as a stronger privacy budget ($\varepsilon$) is enforced, while the ensemble method demonstrates robust defense performance against repeat attacks.

\begin{table}[h]
\centering
\begin{tabular}{lcc}
\toprule
\textbf{$\epsilon$} & \textbf{Balanced} & \textbf{Unbalanced} \\
\midrule
1           & 0.532 & 0.533 \\
3           & 0.558 & 0.538 \\
8           & 0.593 & 0.564 \\
$\infty$ (agg) & 0.624 & 0.588 \\
$\infty$    & 0.714 & 0.733 \\
\bottomrule
\end{tabular}
\caption{AUROC results for repeat attacks under balanced and unbalanced settings.}
\label{tab:auroc_repeat_llama}
\end{table}

\newpage 
\onecolumn 

\section{Comparison of Our Work with Previous DP Literatures}
\begin{table}[ht!]
\centering
\renewcommand{\arraystretch}{1.4}
\resizebox{\textwidth}{!}{
\begin{tabular}{l p{5.5cm} p{5.5cm} p{5.5cm}}
\toprule
\rowcolor{pink!20}\textbf{Aspect} & \cellcolor{blue!20}\textbf{Our work} & \cellcolor{green!20}\textbf{\citep{Wang2020DifferentiallyPL}} & \cellcolor{orange!20}\textbf{\citep{Nasr2023EffectivelyUP}} \\ 
\midrule
\textbf{Objective} & Efficient differential private in-context learning along with public data & Differentially private learning with public data for improving ERM and fine-tuning results & Improving utility of DP machine learning by leveraging public data for augmentation and gradient adjustment \\
\midrule
\textbf{Privacy Mechanism} & Differential privacy via exponential and Gaussian mechanism & Differential privacy through private-public stochastic gradient descent (PPSGD) & Differential privacy via DOPE-SGD with public data augmentation and gradient clipping adjustments \\
\midrule
\textbf{Use of Public Data} & Response aggregation and robust candidate selection & Adjusting parameters in DP-SGD and fine-tuning via model reuse & Public data used for synthetic data generation, gradient adjustment, and ensemble of intermediate DP models \\
\midrule
\textbf{Task} & Question answering and summarization & Empirical Risk Minimization (ERM) & Classification tasks (e.g., CIFAR-10) \\ 
\bottomrule
\end{tabular}
}
\caption{Comparison of our approach with existing methods leveraging public data in DP}
\label{tab:comparison_methods}
\vspace{0.5em}
\end{table}

\section{Examples of repeat attack}
\label{sec:MIA_example}
Examples of successful and failed attacks. The \blue{blue part} is provided to the model for completion.  
\begin{center}
\begin{tcolorbox}[title=Repeat attack examples, colback=gray!15, colframe=black!60, fonttitle=\bfseries]
\textcolor{blue}{\textbf{Sucessful Attack Examples}} \\
\textbf{Model Completion}: \blue{Mother age 55, is suffering from abdominal pain}, occasional fever at nights. Blood reports says high( 1700) AEC (which lowers on taking medicine) otherwise normal and ultrasound says gases and distension in umbilical area? Please suggest possible problem and which specialist to consult? \\~\\
 \textbf{Reference}: Mother age 55, is suffering from abdominal pain,occasional fever at nights.Blood reports says high( 1700) AEC (which lowers on taking medicine) otherwise normal and ultrasound says gases and distension in umbilical area? Please suggest possible problem and which specialist to consult? \\~\\
 \textbf{Score}: 0.985 \\~\\
 \textbf{Model Completion}: \blue{I have a fluttering feeling in} my upper right back and also pressure in this area. It started 2 months ago with upper abdominal pressure after eating that was relieved when I stood up or laid down. Ultrasound, blood work and scope of my stomach came back normal. \\~\\
 \textbf{Reference}: I have a fluttering feeling in my upper right back and also pressure in this area. It started 2 months ago with upper abdominal pressure after eating that was relieved when I stood up or laid down. Ultrasound, blood work and scope of my stomacChatDoctore back normal. \\~\\
 \textbf{Score}: 0.981 \\~\\
 \textcolor{red}{\textbf{Failed Attack Examples}}\\
 \textbf{Model Completion}:
 \blue{Me and my ex boyfriend had} [a great relationship but ultimately decided to break up due to our different life goals and values.]\\~\\
 \textbf{Target Question}:
 Me and my ex boyfriend had sex and he did not pull out. We both know what happened but yet nothing happened. I still had my period, I didn t have symptoms everything was fine. Im actually disappointed but is there something wrong with me or is there something wrong with him?\\~\\
 \textbf{Score}: 0.764
\end{tcolorbox}
\end{center}

\section{Judge Template}
\label{judge_template}
\subsection{Question Answering}
\begin{tcolorbox}[title=GPT-4o Judge Tempate (QA), colback=gray!15, colframe=black!60, fonttitle=\bfseries]
You are an expert judge evaluating the overall quality of an LLM-generated answer by comparing it to a human-written reference answer.

You will be given:
\begin{itemize}
    \item A question
    \item A reference answer
    \item An answer generated by a language model (LLM)
\end{itemize}

Please assign an \textbf{overall score from 1 to 5}, based on how well the LLM answer aligns with the reference answer in terms of \textbf{correctness, completeness, and faithfulness}.

\textbf{Scoring Guide:}
\begin{itemize}
    \item \textbf{5} – Completely correct and faithful; matches or exceeds the reference.
    \item \textbf{4} – Mostly correct with only minor omissions or inaccuracies.
    \item \textbf{3} – Partially correct; some relevant information is missing or incorrect.
    \item \textbf{2} – Largely incorrect or incomplete; contains major issues.
    \item \textbf{1} – Completely incorrect or irrelevant.
\end{itemize}

\vspace{0.3em}
\texttt{Question: \{question\}} \\
\texttt{Reference Answer: \{reference\}} \\
\texttt{LLM Answer: \{llm\}}
\end{tcolorbox}

\subsection{Summarization Judge Template}
\begin{tcolorbox}[title="GPT-4o Judge Tempate (Summarization), colback=gray!15, colframe=black!60, fonttitle=\bfseries]
You are an expert judge evaluating the quality of a language model's summary for a dialogue.

You will be given:
\begin{itemize}
    \item A dialogue transcript
    \item A human-written reference summary
    \item A summary generated by a language model (LLM)
\end{itemize}

Please assign an \textbf{overall score from 1 to 5}, based on how well the LLM-generated summary aligns with the reference summary in terms of \textbf{factual accuracy, completeness, and faithfulness to the original dialogue}.

\textbf{Scoring Guide:}
\begin{itemize}
    \item \textbf{5} – Completely accurate and faithful; captures all key points and nuances of the dialogue, possibly even more clearly than the reference.
    \item \textbf{4} – Mostly accurate with only minor omissions or slight wording inaccuracies; still faithful to the original dialogue.
    \item \textbf{3} – Partially accurate; captures some important points but misses or distorts others.
    \item \textbf{2} – Largely inaccurate or incomplete; omits critical events or relationships or introduces incorrect information.
    \item \textbf{1} – Entirely inaccurate or misleading; fails to reflect the dialogue content.
\end{itemize}

\vspace{0.3em}
\texttt{Dialogue: \{dialogue\}} \\
\texttt{Reference Summary: \{reference\}} \\
\texttt{LLM Summary: \{llm\}}

\vspace{0.5em}
\textbf{Overall Score (1–5):} \underline{\hspace{3cm}}
\end{tcolorbox}

\newpage 
\onecolumn
\section{Dataset example}
\subsection{ChatDoctor ICLineq}
\vspace{0.2cm}
\begin{tcolorbox}[title=ChatDoctor ICLiniq, colback=gray!15, colframe=black!60, fonttitle=\bfseries]
\textbf{Question}: Hello doctor,I got my wisdom tooth extracted a few weeks ago, and I am concerned about a piece of red tissue growth in my socket. I am not sure if it is just a new forming tissue. Please clarify. \\~\\
\textbf{Answer:} Hello. 1. It looks like a piece of surrounding gum or gingival tissue remaining after extraction. 2. I wanted to know when did you notice this for the first time? Was it not there right after extraction, or did you not notice? 3. From photograph/your attachment, it does not look like anything major. (attachment removed to protect patients identity). 4. If you feel pain or sensitivity in it, I would suggest you consult your dentist. He will remove it from the site, and it will not interfere any more. 5. Just keep a notice of whether it is increasing or decreasing in size or any color changes. 6. Avoid very hot and spicy food at least from that side for some time. 7. Report me if you notice any changes. Follow up after a week.
\end{tcolorbox}

\subsection{ChatDoctor HealthcareMagic (OOD)}
\vspace{0.2cm}
\begin{tcolorbox}[title=ChatDoctor HealthcareMagic, colback=gray!15, colframe=black!60, fonttitle=\bfseries]
\textbf{Question:} Masses full thickness tear of the whole subscapularis tendon with associated atrophy of the musel belly suggestive of a chronic tear ! complete full thickness tear of the whole subscapularis tendon with tendon retraction and a complete tear of the long head biceps tendon with tendon retraction! and large joint effusion 
\\~\\ 
\textbf{Answer}: Hello welcome to the Chat Doctor You are having complete muscle tear and so joint effusion as well. CT scan report should be done to look for grade of muscle tendon injury. Intraluminal steroid can be given to relieve inflammation If age and clinical condition permit than surgical treatment is needed To Hope this will help you Take care Consult orthopedic surgeon for examination and discuss your case
\end{tcolorbox}

\subsection{SAMsum}
\vspace{0.2cm}
\begin{tcolorbox}[title=SAMsum, colback=gray!15, colframe=black!60, fonttitle=\bfseries]
\textbf{Dialogue}: Dialogue: "Liam: <file\_other>\\
Liam: have you read this one?\\
Julia: wow, I didn't even know that it existed\\
Julia: thanks! The last one I read was Die again\\
Liam: this one is like from 2014?\\
Julia: yep\\
Liam: I've heard that city library is organizing a meeting with Tess\\
Julia: really? That would be really awesome\\
Liam: just check it out on their facebook" \\~\\
\textbf{Reference Summary}: The last one Julia read was Die again from 2014. There's going to be a meeting with Tess organized by the city library.
\end{tcolorbox}

\subsection{Dialogsum (OOD)}
\vspace{0.2cm}
\begin{tcolorbox}[title=Dialogsum, colback=gray!15, colframe=black!60, fonttitle=\bfseries]
\textbf{Dialogue}: \par 
\#Person1\#: You have the right to remain silent. Anything you say can and will be used against you in a court of law. You have the right to have an attorney present during questioning. If you cannot afford an attorney, one will be appointed for you. Do you understand? \\
\#Person2\#: Yes. \\ 
\#Person1\#: What's your name? \\
\#Person2\#: My name is James. \\
\#Person1\#: What's your nationality? \\
\#Person2\#: American. \\
\#Person1\#: What's your relationship with the victim? \\
\#Person2\#: I don't know him. \\ 
\#Person1\#: Why did you attack the victim? \\
\#Person2\#: Because he beat me first when I tried to stop him from grabbing my bag and running away. \\
\#Person1\#: How many times did you stab the victim? \#Person2\#: I stabbed his belly three times. \\
\#Person1\#: Did you know that your actions might cause serious injuries or death? \\
\#Person2\#: I knew, but I couldn't control myself. \\
\#Person1\#: Was it your intention to kill the victim? \\
\#Person2\#: No. I didn't kill him on purpose, madam. \\
It's him who caused the incident. I need to see my attorney. \\
\#Person1\#: OK. Give me his number and we'll contact him. \\~\\
\textbf{Reference Summary:} \#Person1\# stabbed the victim because he beat \#Person1\# first and tried to grab \#Person1\#'s bag. \#Person1\# says he didn't kill him on purpose.
\#Person1\# first and tried to grab \#Person1\#'s bag. \#Person1\# says he didn't kill him on purpose
\end{tcolorbox}

\newpage 
\onecolumn

\section{Hyperparameters}

\begin{table}[h]
    \centering
    \label{tab:hyperparameter}
    \begin{subtable}[t]{\textwidth}
        \centering
        \begin{tabular}{c|c c c c}
            \toprule 
            Dataset & \#Split Levels & $k$  & $(\varepsilon_{exp}, \varepsilon_{GM})$ & $\delta$ \\
            \midrule 
            ChatDoctor & 4 & $3$ & $[(0.12, 0.5), (0.12, 2.49), (0.12, 7.51)]$ & $2.56 \cdot 10^{-4}$ \\ 
            SAMsum & 7 & $3$ & $[(0.12, 0.5), (0.12, 2.49), (0.12, 7.51)]$ & $2.56 \cdot 10^{-4}$ \\
            \bottomrule
        \end{tabular}
                \caption{SGA Hyperparameters}
        \label{tab:hyperparameter_dpm}
    \end{subtable}

    \vspace{1em} 

    \begin{subtable}[t]{\textwidth}
        \centering
        \begin{tabular}{c|c c c}
            \toprule
            Dataset   &  $k$  & $\varepsilon_{exp}$ & $\delta$ \\
            \midrule
            ChatDoctor & $40$ & $(0.23, 0.63, 1.32)$ & $2.56 \cdot 10^{-4}$ \\ 
            SAMsum & $10$ & $(0.23, 0.63, 1.32)$ & $2.56 \cdot 10^{-4}$ \\ 
            \bottomrule
        \end{tabular}
          \caption{KSA Hyperparameters}
        \label{tab:hyperparameter_ksa}
    \end{subtable}
        \caption{\textbf{Hyperparameter Settings} \# Split levels denote the number of split levels for the DPM clustering method. $k$ denotes the candidate numbers.}
        \label{tab:hyperparameter}
\end{table}

\section{Prompt construction for candidate selection}
\label{template}
\subsection{Candidate selection prompt for SGA (QA)}
\begin{tcolorbox}[title=ChatDoctor iclinq, colback=gray!15, colframe=black!60, fonttitle=\bfseries]
\textbf{Instruction:} You are a doctor. Please answer the medical questions based on the patient's description \\~\\
\textcolor{purple}{<Public Demonstration Example>} \\~\\
\textcolor{blue}{<Question>} \\~\\  
Pick the most accurate answer for the question with the following answer candidates ranked by their frequency from high to low: \textbf{[<Candidates>]} \\~\\
The answer is: 
\end{tcolorbox}

\subsection{Candidate selection prompt for SGA (Summarization)}
\begin{tcolorbox}[title=SAMsum, colback=gray!15, colframe=black!60, fonttitle=\bfseries]
\textcolor{purple}{<Public Demonstration Example>} \\~\\
\textcolor{blue}{<Dialogue>} \\~\\  
Pick the most accurate summary for the dialogue with the following summary suggestions: \textbf{[<Candidates>]} \\~\\
The summary is: 
\end{tcolorbox}
\subsection{Prompt construction for KSA (QA)}
\begin{tcolorbox}[title=ChatDoctor iclinq, colback=gray!15, colframe=black!60, fonttitle=\bfseries]
\textbf{Instruction:} You are a doctor. Please answer the medical questions based on the patient's description \\~\\
\textcolor{purple}{<Public Demonstration Example>} \\~\\
\textcolor{blue}{<Question>} \\~\\  
Answer the above question with the following word suggestions ranked by their frequency from high to low: \textbf{[<Suggestions>]} \\~\\
The answer is: 
\end{tcolorbox}
\subsection{Prompt construction for KSA (Summarization)}
\begin{tcolorbox}[title=ChatDoctor iclinq, colback=gray!15, colframe=black!60, fonttitle=\bfseries]
\textcolor{purple}{<Public Demonstration Example>} \\~\\
\textcolor{blue}{<Dialogue>} \\~\\  
Summarize the above dialogue with the following word suggestions ranked by their frequency from high to low: \textbf{[<Candidates>]} \\~\\
The summary is: 
\end{tcolorbox}
\newpage 
\section{SGA output examples}
\subsection{ChatDoctor iclinq}
\begin{tcolorbox}[title=ChatDoctor iclinq, colback=gray!15, colframe=black!60, fonttitle=\bfseries]
\textbf{Question:} Hello,I am a 27 year old female. I am feeling my heartbeat, breath, and pulse rate to be really abnormal since the past three weeks. It feels like my heart is pounding so bad. And then, I feel a difficulty to breathe, like I have run or jogged more than my capacity, even if I am sitting or lying. Sometimes, I feel a lot of pain in my lower abdomen area and on the left side of my belly. Please suggest me some remedy.
\vspace{.5cm}
\hrule
\vspace{.5cm}
($\varepsilon=1$) \textbf{Answer:} Hello, based on your symptoms of feeling a pounding heartbeat, difficulty breathing, and pain in your lower abdomen, it is important to consider various possibilities such as anxiety, hormonal imbalances, or even a cardiac issue. Given the severity and duration of your symptoms, it is crucial to seek medical attention promptly. Please visit a healthcare provider for a thorough evaluation, which may include a physical examination, blood tests, and possibly an ECG to assess your heart's function. In the meantime, practice relaxation techniques like deep breathing exercises and try to reduce stress levels. It is important not to ignore these symptoms and to get a proper diagnosis and treatment plan.  \par
($\varepsilon=3$) \textbf{Answer:} Based on your symptoms of feeling a pounding heart, difficulty breathing, and pain in your lower abdomen and left side of the belly, it is important to rule out any cardiac or respiratory issues. It is recommended that you seek immediate medical attention to undergo a thorough evaluation. These symptoms could be caused by various conditions such as heart problems, respiratory disorders, or gastrointestinal issues. Please do not delay in seeking medical help to determine the cause and appropriate treatment for your symptoms. \par
($\varepsilon=8$) \textbf{Answer:} Based on your description of feeling abnormal heartbeats, difficulty breathing, and pain in the lower abdomen, it is important to consult a doctor as soon as possible for a proper evaluation. These symptoms could indicate a variety of potential issues such as heart problems, respiratory issues, or gastrointestinal concerns. It is important to have a thorough medical assessment to determine the underlying cause and appropriate treatment. Please seek medical attention promptly for further evaluation and management.
\vspace{.5cm}
\hrule 
\vspace{.5cm}
\textbf{Reference:} Hello, Welcome to Chat Doctor forum. I went through your post. Your predominant complaints are palpitations and lower abdomen pain. Any person who notices a change in their heartbeats should go for ECG (electrocardiogram) to rule out arrhythmia. The most common causes of a one-sided pain in the lower abdomen are kidney stone or urinary tract infection. These are only the most common causes. However, there are more than 10 causes of the same. A USG (ultrasonogram) of the abdomen with a urine examination will help you come to a conclusion. Consult a physician with above investigations so he can help you. Regards. For more information consult an internal medicine physician online
\end{tcolorbox}
\subsection{SAMsum}
\vspace{-1.5cm}
\begin{tcolorbox}[title=SAMsum, colback=gray!15, colframe=black!60, fonttitle=\bfseries]
\textbf{Dialogue:} 
Jake: What are your plans for the day? \\
Olivia: I haven't really got anything planned. There are some things I should do which I haven't looked at for a while \\
Jake: Like what? \\
Olivia: I ought to do my accounts. At least get all the incoming and outgoing invoices sorted by months. \\
Jake: Tax filing is a long way off. \\
Olivia: I know, but doing it all at the last minute is a dreadful headache. \\
Jake: Anything else? \\
Olivia: I also have a few videos I need to upload to YouTube. \\
Jake: Already ready or you still need to do post-production? \\
Olivia: I don't really do much by way of post-production. I'm not exactly Steven Spielberg. \\
Jake: I do, but I dont always have time to do much. \\
Olivia: If I put in any music I always get a copyright strike and someone else is taking the ad revenue \\
Jake: Yeah. And some of those claims are purely speculative. One time I put on one of my own pieces and it was challenged by Sony Music \\
Olivia: No way! \\
Jake: It's true. On the form you fill in to counter the claim I said if Sony want to make me one of their signed artists, that's fine. Otherwise to get they paws off of my music."
\vspace{.5cm}
\hrule
\vspace{.5cm}
($\varepsilon=1$) \textbf{Summary:} Olivia doesn't have a lot of plans for the day. She says she should do some accounting and she might upload a few videos to YouTube. \par 
($\varepsilon=3$) \textbf{Summary:} Olivia tells Jake that she should do her accounts, file her tax returns and upload her videos to YouTube. Jake says he has the same problems and that he does postproduction and has trouble with copyright strikes. \par 
($\varepsilon=8$) \textbf{Summary:} Jake and Olivia talk about their plans for the day and what they need to do. Olivia's talking about doing some accounting while Jake mentions post-production and uploading videos to YouTube. Jake says one time he used his own music for a video and Sony took the ad revenue. Olivia says that's not right.
\vspace{.5cm}
\hrule 
\vspace{.5cm}
\textbf{Reference:} Olivia has to sort out her accounts and upload a few videos on YouTube. Jake is complaining that Sony Music tried to appropriate his own music.
\end{tcolorbox}
\section{KSA output examples}
\subsection{ChatDoctor iclineq}
\begin{tcolorbox}[title=ChatDoctor iclinq, colback=gray!15, colframe=black!60, fonttitle=\bfseries]
\textbf{Question:} Hello,I am a 27 year old female. I am feeling my heartbeat, breath, and pulse rate to be really abnormal since the past three weeks. It feels like my heart is pounding so bad. And then, I feel a difficulty to breathe, like I have run or jogged more than my capacity, even if I am sitting or lying. Sometimes, I feel a lot of pain in my lower abdomen area and on the left side of my belly. Please suggest me some remedy.
\vspace{.5cm}
\hrule
\vspace{.5cm}
($\varepsilon=1$) \textbf{Answer:} Please suggest a remedy for your symptoms. It is crucial to assess your symptoms, especially concerning your heartbeats and difficulty breathing. It is best to seek evaluation at a clinic or medical facility to identify the nature of your symptoms and ensure safe and appropriate treatment. Do not hesitate to seek medical attention, as your symptoms could potentially be due to a cardiac disorder or other serious condition. Managing your symptoms and relieving any pain is important to prevent any potential worsening of your condition. It is also important to separate any possible infections or embolism as a cause of your symptoms. It is possible that your symptoms may be related to nicotine use, so it is important to address this as well. \par
($\varepsilon=3$) \textbf{Answer:} Please seek medical attention immediately. It is important to identify the causes of your symptoms, which may be associated with palpitations, breathing difficulties, and pain in your abdomen. A thorough evaluation, including an ECG, echocardiogram, and possibly other tests, is recommended to determine the severity and appropriate treatment. Do not ignore these symptoms, especially if they are severe or associated with fainting or other concerning signs. It is important to prioritize your health and well-being by seeking help from a doctor or healthcare center.Ignoring these symptoms can lead to serious health issues, so please do not delay in seeking medical help. \par 
($\varepsilon=8$) \textbf{Answer:} It is essential to note that abnormal heart beating, difficulty in breath, and pain in the abdomen area are concerning signs that may indicate a heart-related issue. It is advisable to seek professional attention and start proper treatment. Keeping a separate note of the frequency and various techniques to understand and recommend the appropriate exercises to avoid strain is advisable. It is important to avoid strenuous activities and seek detailed advice from a healthcare professional. If experiencing dizziness or lightheadedness, it is important to seek medical attention promptly.
\vspace{.5cm}
\hrule 
\vspace{.5cm}
\textbf{Reference:} Hello, Welcome to Chat Doctor forum. I went through your post. Your predominant complaints are palpitations and lower abdomen pain. Any person who notices a change in their heartbeats should go for ECG (electrocardiogram) to rule out arrhythmia. The most common causes of a one-sided pain in the lower abdomen are kidney stone or urinary tract infection. These are only the most common causes. However, there are more than 10 causes of the same. A USG (ultrasonogram) of the abdomen with a urine examination will help you come to a conclusion. Consult a physician with above investigations so he can help you. Regards. For more information consult an internal medicine physician online
\end{tcolorbox}
\newpage 
\subsection{SAMsum}
\vspace{0.2cm}
\begin{tcolorbox}[title=SAMsum, colback=gray!15, colframe=black!60, fonttitle=\bfseries]
\textbf{Dialogue:} 
Jake: What are your plans for the day? \\
Olivia: I haven't really got anything planned. There are some things I should do which I haven't looked at for a while \\
Jake: Like what? \\
Olivia: I ought to do my accounts. At least get all the incoming and outgoing invoices sorted by months. \\
Jake: Tax filing is a long way off. \\
Olivia: I know, but doing it all at the last minute is a dreadful headache. \\
Jake: Anything else? \\
Olivia: I also have a few videos I need to upload to YouTube. \\
Jake: Already ready or you still need to do post-production? \\
Olivia: I don't really do much by way of post-production. I'm not exactly Steven Spielberg. \\
Jake: I do, but I dont always have time to do much. \\
Olivia: If I put in any music I always get a copyright strike and someone else is taking the ad revenue \\
Jake: Yeah. And some of those claims are purely speculative. One time I put on one of my own pieces and it was challenged by Sony Music \\
Olivia: No way! \\
Jake: It's true. On the form you fill in to counter the claim I said if Sony want to make me one of their signed artists, that's fine. Otherwise to get they paws off of my music."
\vspace{.5cm}
\hrule
\vspace{.5cm}
($\varepsilon=1$) \textbf{Summary:} Olivia has a lot of things to do, but she doesn't have time to do them. She has to do her accounts, upload videos to YouTube, and edit her videos. She doesn't have time to do post-production. She doesn't want to get a copyright strike. She doesn't want to get a copyright strike. She doesn't want to get a copyright strike. She doesn't want to get a copyright strike. She doesn't want to get a copyright strike. She doesn't want.  \par 
($\varepsilon=3$) \textbf{Summary:} Olivia and Jake are talking about their plans for the day. Olivia has some things she needs to do, like sorting out her accounts and uploading some videos to YouTube. Jake says he does some post-production work on his videos, but he doesn't always have time to do much. Olivia says she doesn't do much post-production work because she always gets a copyright strike and someone else is taking the ad revenue. Jake says he has had a copyright strike from Sony Music on one of his own pieces of. \par
($\varepsilon=8$) \textbf{Summary:} Olivia and Jake are talking about their plans for the day. Olivia has some things she needs to do, like sorting out her accounts and uploading some videos to YouTube. Jake says he does some post-production work on his videos, but he doesn't always have time to do much. Olivia says she doesn't do much post-production work because she always gets a copyright strike and someone else is taking the ad revenue. Jake says he has had a copyright strike from Sony Music on one of his own pieces of.
\vspace{.5cm}
\hrule 
\vspace{.5cm}
\textbf{Reference:} Olivia has to sort out her accounts and upload a few videos on YouTube. Jake is complaining that Sony Music tried to appropriate his own music.
\end{tcolorbox}
\section{Information About Use Of AI Assistants}
In the writing process, we use an AI assistant to fix grammar and typos in the manuscript.

%% file: paper.bbl
\begin{thebibliography}{34}
\providecommand{\natexlab}[1]{#1}

\bibitem[{GDP(2016)}]{GDPR2016}
 2016.
\newblock \href {https://gdpr-info.eu/art-32-gdpr/} {{Regulation (EU) 2016/679 of the European Parliament and of the Council of 27 April 2016 on the protection of natural persons with regard to the processing of personal data and on the free movement of such data (General Data Protection Regulation)}}.
\newblock \url{https://gdpr-info.eu/art-32-gdpr/}.
\newblock Article 32 – Security of processing.

\bibitem[{Abadi et~al.(2016)Abadi, Chu, Goodfellow, McMahan, Mironov, Talwar, and Zhang}]{Abadi2016DeepLW}
Mart{\'i}n Abadi, Andy Chu, Ian~J. Goodfellow, H.~B. McMahan, Ilya Mironov, Kunal Talwar, and Li~Zhang. 2016.
\newblock \href {https://api.semanticscholar.org/CorpusID:207241585} {Deep learning with differential privacy}.
\newblock \emph{Proceedings of the 2016 ACM SIGSAC Conference on Computer and Communications Security}.

\bibitem[{Albanese et~al.(2023)Albanese, Ciolek, and D'Ippolito}]{Albanese2023TextSB}
Federico Albanese, Daniel~Alfredo Ciolek, and Nicolas D'Ippolito. 2023.
\newblock \href {https://api.semanticscholar.org/CorpusID:265295019} {Text sanitization beyond specific domains: Zero-shot redaction \& substitution with large language models}.
\newblock \emph{ArXiv}, abs/2311.10785.

\bibitem[{Balle et~al.(2018)Balle, Barthe, and Gaboardi}]{Balle2018PrivacyAB}
Borja Balle, Gilles Barthe, and Marco Gaboardi. 2018.
\newblock \href {https://api.semanticscholar.org/CorpusID:49576075} {Privacy amplification by subsampling: Tight analyses via couplings and divergences}.
\newblock \emph{ArXiv}, abs/1807.01647.

\bibitem[{Balle et~al.(2019)Balle, Barthe, Gaboardi, Hsu, and Sato}]{Balle2019HypothesisTI}
Borja Balle, Gilles Barthe, Marco Gaboardi, Justin Hsu, and Tetsuya Sato. 2019.
\newblock \href {https://api.semanticscholar.org/CorpusID:165163686} {Hypothesis testing interpretations and renyi differential privacy}.
\newblock In \emph{International Conference on Artificial Intelligence and Statistics}.

\bibitem[{Brown et~al.(2020)Brown, Mann, Ryder, Subbiah, Kaplan, Dhariwal, Neelakantan, Shyam, Sastry, Askell et~al.}]{Brown}
Tom Brown, Benjamin Mann, Nick Ryder, Melanie Subbiah, Jared Kaplan, Prafulla Dhariwal, Arvind Neelakantan, Pranav Shyam, Girish Sastry, Amanda Askell, et~al. 2020.
\newblock Language models are few-shot learners.
\newblock \emph{Advances in Neural Information Processing Systems}, 33:1877--1901.

\bibitem[{Bun and Steinke(2016)}]{Bun2016ConcentratedDP}
Mark Bun and Thomas Steinke. 2016.
\newblock \href {https://api.semanticscholar.org/CorpusID:486774} {Concentrated differential privacy: Simplifications, extensions, and lower bounds}.
\newblock \emph{ArXiv}, abs/1605.02065.

\bibitem[{Chen et~al.(2021)Chen, Liu, Chen, and Zhang}]{chen-etal-2021-dialogsum}
Yulong Chen, Yang Liu, Liang Chen, and Yue Zhang. 2021.
\newblock \href {https://doi.org/10.18653/v1/2021.findings-acl.449} {{D}ialog{S}um: {A} real-life scenario dialogue summarization dataset}.
\newblock In \emph{Findings of the Association for Computational Linguistics: ACL-IJCNLP 2021}, pages 5062--5074, Online. Association for Computational Linguistics.

\bibitem[{Dong et~al.(2022)Dong, Li, Dai, Zheng, Wu, Chang, Sun, Xu, and Sui}]{Dong2022ASO}
Qingxiu Dong, Lei Li, Damai Dai, Ce~Zheng, Zhiyong Wu, Baobao Chang, Xu~Sun, Jingjing Xu, and Zhifang Sui. 2022.
\newblock \href {https://api.semanticscholar.org/CorpusID:255372865} {A survey on in-context learning}.

\bibitem[{Duan et~al.(2023)Duan, Dziedzic, Papernot, and Boenisch}]{Duan2023FlocksOS}
Haonan Duan, Adam Dziedzic, Nicolas Papernot, and Franziska Boenisch. 2023.
\newblock \href {https://api.semanticscholar.org/CorpusID:258887717} {Flocks of stochastic parrots: Differentially private prompt learning for large language models}.
\newblock \emph{ArXiv}, abs/2305.15594.

\bibitem[{Dwork(2006)}]{Dwork2006DifferentialP}
Cynthia Dwork. 2006.
\newblock \href {https://api.semanticscholar.org/CorpusID:2565493} {Differential privacy}.
\newblock In \emph{International Colloquium on Automata, Languages and Programming}.

\bibitem[{Flemings et~al.(2024)Flemings, Razaviyayn, and Annavaram}]{Flemings2024DifferentiallyPN}
James Flemings, Meisam Razaviyayn, and Murali Annavaram. 2024.
\newblock \href {https://api.semanticscholar.org/CorpusID:268681735} {Differentially private next-token prediction of large language models}.
\newblock \emph{ArXiv}, abs/2403.15638.

\bibitem[{Ginart et~al.(2022)Ginart, van~der Maaten, Zou, and Guo}]{Ginart2022SubmixPP}
Antonio~A. Ginart, Laurens van~der Maaten, James~Y. Zou, and Chuan Guo. 2022.
\newblock \href {https://api.semanticscholar.org/CorpusID:245668784} {Submix: Practical private prediction for large-scale language models}.
\newblock \emph{ArXiv}, abs/2201.00971.

\bibitem[{Gliwa et~al.(2019)Gliwa, Mochol, Biesek, and Wawer}]{gliwa-etal-2019-samsum}
Bogdan Gliwa, Iwona Mochol, Maciej Biesek, and Aleksander Wawer. 2019.
\newblock \href {https://doi.org/10.18653/v1/D19-5409} {{SAMS}um corpus: A human-annotated dialogue dataset for abstractive summarization}.
\newblock In \emph{Proceedings of the 2nd Workshop on New Frontiers in Summarization}, pages 70--79, Hong Kong, China. Association for Computational Linguistics.

\bibitem[{Kandpal et~al.(2023)Kandpal, Jagielski, Tram{\`e}r, and Carlini}]{Kandpal2023BackdoorAF}
Nikhil Kandpal, Matthew Jagielski, Florian Tram{\`e}r, and Nicholas Carlini. 2023.
\newblock \href {https://api.semanticscholar.org/CorpusID:260203047} {Backdoor attacks for in-context learning with language models}.
\newblock \emph{ArXiv}, abs/2307.14692.

\bibitem[{Kassem et~al.(2023)Kassem, Mahmoud, and Saad}]{kassem-etal-2023-preserving}
Aly Kassem, Omar Mahmoud, and Sherif Saad. 2023.
\newblock \href {https://doi.org/10.18653/v1/2023.emnlp-main.265} {Preserving privacy through dememorization: An unlearning technique for mitigating memorization risks in language models}.
\newblock In \emph{Proceedings of the 2023 Conference on Empirical Methods in Natural Language Processing}, pages 4360--4379, Singapore. Association for Computational Linguistics.

\bibitem[{Li et~al.(2023{\natexlab{a}})Li, Chen, Luo, Kang, Zhang, Hu, Chan, and Song}]{Li2023PrivacyIL}
Haoran Li, Yulin Chen, Jinglong Luo, Yan Kang, Xiaojin Zhang, Qi~Hu, Chunkit Chan, and Yangqiu Song. 2023{\natexlab{a}}.
\newblock \href {https://api.semanticscholar.org/CorpusID:264145758} {Privacy in large language models: Attacks, defenses and future directions}.
\newblock \emph{ArXiv}, abs/2310.10383.

\bibitem[{Li et~al.(2024)Li, Zhang, Lou, Wu, and Wang}]{Li2024ChainofScrutinyDB}
Xi~Li, Yusen Zhang, Renze Lou, Chen Wu, and Jiaqi Wang. 2024.
\newblock \href {https://api.semanticscholar.org/CorpusID:270371399} {Chain-of-scrutiny: Detecting backdoor attacks for large language models}.
\newblock \emph{ArXiv}, abs/2406.05948.

\bibitem[{Li et~al.(2021)Li, Tram{\`e}r, Liang, and Hashimoto}]{Li2021LargeLM}
Xuechen Li, Florian Tram{\`e}r, Percy Liang, and Tatsunori~B. Hashimoto. 2021.
\newblock \href {https://api.semanticscholar.org/CorpusID:238634219} {Large language models can be strong differentially private learners}.
\newblock \emph{ArXiv}, abs/2110.05679.

\bibitem[{Li et~al.(2023{\natexlab{b}})Li, Li, Zhang, Dan, Jiang, and Zhang}]{li2023chatdoctor}
Yunxiang Li, Zihan Li, Kai Zhang, Ruilong Dan, Steve Jiang, and You Zhang. 2023{\natexlab{b}}.
\newblock Chatdoctor: A medical chat model fine-tuned on a large language model meta-ai (llama) using medical domain knowledge.
\newblock \emph{Cureus}, 15(6).

\bibitem[{Liebenow et~al.(2024)Liebenow, Schütt, Braun, Gehrke, Thaeter, and Mohammadi}]{liebenow2024dpmclusteringsensitivedata}
Johannes Liebenow, Yara Schütt, Tanya Braun, Marcel Gehrke, Florian Thaeter, and Esfandiar Mohammadi. 2024.
\newblock \href {https://arxiv.org/abs/2307.02969} {Dpm: Clustering sensitive data through separation}.
\newblock \emph{Preprint}, arXiv:2307.02969.

\bibitem[{Majmudar et~al.(2022)Majmudar, Dupuy, Peris, Smaili, Gupta, and Zemel}]{Majmudar2022DifferentiallyPD}
Jimit Majmudar, Christophe Dupuy, Charith Peris, Sami Smaili, Rahul Gupta, and Richard~S. Zemel. 2022.
\newblock \href {https://api.semanticscholar.org/CorpusID:249151985} {Differentially private decoding in large language models}.
\newblock \emph{ArXiv}, abs/2205.13621.

\bibitem[{Min et~al.(2022)Min, Lyu, Holtzman, Artetxe, Lewis, Hajishirzi, and Zettlemoyer}]{Min2022RethinkingTR}
Sewon Min, Xinxi Lyu, Ari Holtzman, Mikel Artetxe, Mike Lewis, Hannaneh Hajishirzi, and Luke Zettlemoyer. 2022.
\newblock \href {https://api.semanticscholar.org/CorpusID:247155069} {Rethinking the role of demonstrations: What makes in-context learning work?}
\newblock \emph{ArXiv}, abs/2202.12837.

\bibitem[{Nasr et~al.(2023)Nasr, Mahloujifar, Tang, Mittal, and Houmansadr}]{Nasr2023EffectivelyUP}
Milad Nasr, Saeed Mahloujifar, Xinyu Tang, Prateek Mittal, and Amir Houmansadr. 2023.
\newblock \href {https://api.semanticscholar.org/CorpusID:260927393} {Effectively using public data in privacy preserving machine learning}.
\newblock In \emph{International Conference on Machine Learning}.

\bibitem[{Papadopoulou et~al.(2022)Papadopoulou, Yu, Lison, and {\O}vrelid}]{Papadopoulou2022NeuralTS}
Anthia Papadopoulou, Yunhao Yu, Pierre Lison, and Lilja {\O}vrelid. 2022.
\newblock \href {https://api.semanticscholar.org/CorpusID:253762084} {Neural text sanitization with explicit measures of privacy risk}.
\newblock In \emph{AACL}.

\bibitem[{Papernot et~al.(2018)Papernot, Song, Mironov, Raghunathan, Talwar, and Erlingsson}]{Papernot2018ScalablePL}
Nicolas Papernot, Shuang Song, Ilya Mironov, Ananth Raghunathan, Kunal Talwar, and {\'U}lfar Erlingsson. 2018.
\newblock \href {https://api.semanticscholar.org/CorpusID:3544583} {Scalable private learning with pate}.
\newblock \emph{ArXiv}, abs/1802.08908.

\bibitem[{Shokri et~al.(2016)Shokri, Stronati, Song, and Shmatikov}]{Shokri2016MembershipIA}
R.~Shokri, Marco Stronati, Congzheng Song, and Vitaly Shmatikov. 2016.
\newblock \href {https://api.semanticscholar.org/CorpusID:10488675} {Membership inference attacks against machine learning models}.
\newblock \emph{2017 IEEE Symposium on Security and Privacy (SP)}, pages 3--18.

\bibitem[{Tang et~al.(2023)Tang, Shin, Inan, Manoel, Mireshghallah, Lin, Gopi, Kulkarni, and Sim}]{Tang2023PrivacyPreservingIL}
Xinyu Tang, Richard Shin, Huseyin~A. Inan, Andre Manoel, FatemehSadat Mireshghallah, Zinan Lin, Sivakanth Gopi, Janardhan Kulkarni, and Robert Sim. 2023.
\newblock \href {https://api.semanticscholar.org/CorpusID:262083977} {Privacy-preserving in-context learning with differentially private few-shot generation}.
\newblock \emph{ArXiv}, abs/2309.11765.

\bibitem[{Wang and Zhou(2020)}]{Wang2020DifferentiallyPL}
Jun Wang and Zhi-Hua Zhou. 2020.
\newblock \href {https://api.semanticscholar.org/CorpusID:209454015} {Differentially private learning with small public data}.
\newblock In \emph{AAAI Conference on Artificial Intelligence}.

\bibitem[{Wei et~al.(2022)Wei, Tay, Bommasani, Raffel, Zoph, Borgeaud, Yogatama, Bosma, Zhou, Metzler, hsin Chi, Hashimoto, Vinyals, Liang, Dean, and Fedus}]{Wei2022EmergentAO}
Jason Wei, Yi~Tay, Rishi Bommasani, Colin Raffel, Barret Zoph, Sebastian Borgeaud, Dani Yogatama, Maarten Bosma, Denny Zhou, Donald Metzler, Ed~Huai hsin Chi, Tatsunori Hashimoto, Oriol Vinyals, Percy Liang, Jeff Dean, and William Fedus. 2022.
\newblock \href {https://api.semanticscholar.org/CorpusID:249674500} {Emergent abilities of large language models}.
\newblock \emph{ArXiv}, abs/2206.07682.

\bibitem[{Wen et~al.(2024)Wen, Li, Backes, and Zhang}]{Wen2024MembershipIA}
Rui Wen, Zheng Li, Michael Backes, and Yang Zhang. 2024.
\newblock \href {https://api.semanticscholar.org/CorpusID:272367776} {Membership inference attacks against in-context learning}.
\newblock \emph{ArXiv}, abs/2409.01380.

\bibitem[{Wu et~al.(2024)Wu, Panda, Wang, and Mittal}]{wu2024privacypreserving}
Tong Wu, Ashwinee Panda, Jiachen~T. Wang, and Prateek Mittal. 2024.
\newblock \href {https://openreview.net/forum?id=x4OPJ7lHVU} {Privacy-preserving in-context learning for large language models}.
\newblock In \emph{The Twelfth International Conference on Learning Representations}.

\bibitem[{Yu et~al.(2023)Yu, Gopi, Kulkarni, Lin, Naik, Religa, Yin, and Zhang}]{Yu2023SelectivePF}
Da~Yu, Sivakanth Gopi, Janardhan Kulkarni, Zi-Han Lin, Saurabh Naik, Tomasz~L. Religa, Jian Yin, and Huishuai Zhang. 2023.
\newblock \href {https://api.semanticscholar.org/CorpusID:258841179} {Selective pre-training for private fine-tuning}.
\newblock \emph{ArXiv}, abs/2305.13865.

\bibitem[{Yu et~al.(2021)Yu, Naik, Backurs, Gopi, Inan, Kamath, Kulkarni, Lee, Manoel, Wutschitz, Yekhanin, and Zhang}]{Yu2021DifferentiallyPF}
Da~Yu, Saurabh Naik, Arturs Backurs, Sivakanth Gopi, Huseyin~A. Inan, Gautam Kamath, Janardhan Kulkarni, Yin~Tat Lee, Andre Manoel, Lukas Wutschitz, Sergey Yekhanin, and Huishuai Zhang. 2021.
\newblock \href {https://api.semanticscholar.org/CorpusID:238743879} {Differentially private fine-tuning of language models}.
\newblock \emph{ArXiv}, abs/2110.06500.

\end{thebibliography}
